%% file: TMLR.tex
\documentclass[10pt]{article} 
\usepackage[preprint]{tmlr}

\usepackage{hyperref}
\usepackage{url}



\usepackage{algorithm}
\usepackage{algorithmic}

\usepackage[draft]{graphicx}

\usepackage{enumitem}

\usepackage{microtype}
\usepackage{subfigure}
\usepackage{booktabs}
\usepackage{comment}
\usepackage{wrapfig}
\usepackage{arydshln}
\usepackage{enumitem}
\usepackage{multirow}
\usepackage{url}
\usepackage{natbib}
\usepackage{authblk}

\usepackage{amsmath}
\usepackage{amssymb}
\usepackage{mathtools}
\usepackage{amsfonts}
\usepackage{bm}
\usepackage{bbm}

\input{Macros/math_PMLR.tex}
\input{Macros/theorems.tex}

\usepackage[capitalize,noabbrev]{cleveref}

\allowdisplaybreaks

\title{Semi-Supervised Treatment Effect Estimation with Unlabeled Covariates for Prediction-Powered Causal Inference}

\author{\name Masahiro Kato \email mkato-csecon@g.ecc.u-tokyo.ac.jp \\
      The University of Tokyo\\
      Osaka Metropolitan University}

\def\month{MM}  
\def\year{YYYY} 
\def\openreview{\url{https://openreview.net/forum?id=XXXX}} 

\begin{document}

\maketitle

\input{Head/Abstract}


\input{Main}

\end{document}

%% file: Macros/math_PMLR.tex
\def\eqref#1{equation~\ref{#1}}

\def\1{\bm{1}}

\def\rmd{{\mathrm{d}}}

\def\rmp{{\mathrm{p}}}

\DeclareMathAlphabet{\mathsfit}{\encodingdefault}{\sfdefault}{m}{sl}
\SetMathAlphabet{\mathsfit}{bold}{\encodingdefault}{\sfdefault}{bx}{n}

\def\calA{{\mathcal{A}}}

\def\calD{{\mathcal{D}}}

\def\calI{{\mathcal{I}}}

\def\calL{{\mathcal{L}}}
\def\calM{{\mathcal{M}}}
\def\calN{{\mathcal{N}}}

\def\calP{{\mathcal{P}}}

\def\calT{{\mathcal{T}}}

\def\calX{{\mathcal{X}}}
\def\calY{{\mathcal{Y}}}

\def\bbE{{\mathbb{E}}}

\def\bbR{{\mathbb{R}}}

\newcommand{\p}[1]{\left(#1\right)}
\newcommand{\sqb}[1]{\left[#1\right]}
\newcommand{\cb}[1]{\left\{#1\right\}}

\newcommand{\Bigp}[1]{\Big(#1\Big)}
\newcommand{\Bigsqb}[1]{\Big[#1\Big]}

\newcommand{\Biggp}[1]{\Bigg(#1\Bigg)}
\newcommand{\Biggsqb}[1]{\Bigg[#1\Bigg]}

\newcommand{\bigp}[1]{\big(#1\big)}
\newcommand{\bigsqb}[1]{\big[#1\big]}

\newcommand{\abs}[1]{\left|#1\right|}

\newcommand{\indep}{\rotatebox[origin=c]{90}{$\models$}}

\newcommand{\iid}{ \stackrel{\mathrm{i.i.d}}{\sim} }
\newcommand{\annot}[2]{\underbrace{#1}_{\text{#2}}}

\newcommand{\Exp}[1]{\mathbb{E}\left[#1\right]}

\newcommand{\BigExp}[1]{\mathbb{E}\Big[#1\Big]}

\DeclareMathOperator*{\argmin}{arg\,min}

%% file: Macros/theorems.tex
\usepackage{amsthm}

\theoremstyle{plain}

\newtheorem{theorem}{Theorem}[section]
\newtheorem{lemma}[theorem]{Lemma}

\newtheorem{corollary}[theorem]{Corollary}
\newtheorem{proposition}[theorem]{Proposition}

\newtheorem{assumption}{Assumption}[section]

\newtheorem*{remark}{Remark}

%% file: Head/Abstract.tex
\begin{abstract}
    This study investigates treatment effect estimation in the semi-supervised setting, also can be interpreted as prediction-powered inference. In our setting, we can use not only the standard triple of covariates, treatment indicator, and outcome, but also unlabeled auxiliary covariates. For this problem, we develop efficiency bounds and efficient estimators whose asymptotic variance aligns with the efficiency bound. In the analysis, we introduce two different data-generating processes: the one-sample setting and the two-sample setting. The one-sample setting considers the case where we can observe treatment indicators and outcomes for a part of the dataset, which is also called the censoring setting. In contrast, the two-sample setting considers two independent datasets with labeled and unlabeled data, which is also called the case-control setting or the stratified setting. In both settings, we find that by incorporating auxiliary covariates, we can lower the efficiency bound and obtain an estimator with an asymptotic variance smaller than that without such auxiliary covariates. We frame our framework as prediction-powered causal inference. 
\end{abstract}

%% file: Main.tex
\section{Introduction}
A core interest in causal inference is estimating treatment effects, including the average treatment effect \citep[ATE,][]{Imbens2015causalinference}. In the standard setup, we estimate such treatment effects from triples of covariates, a treatment indicator, and outcomes. As in other statistical analyses, accuracy depends not only on the statistical method but also on the amount and type of data available. While randomized controlled trials are the gold standard, they are often infeasible. Therefore, in many practical scenarios, we use observational data to perform causal inference. However, observational data are also not necessarily easy to collect. In particular, treatment variables and the corresponding outcomes are often costly, whereas covariates are usually easy to gather.

Under this practical scenario, we consider estimating ATEs more accurately using auxiliary unlabeled covariates, even when treatment variables and outcomes are missing. We also discuss the average treatment effect on the treated, because it is often the causal target in observational studies where treatment participation itself determines the relevant population. This setting corresponds to semi-supervised learning in machine learning, where we utilize both labeled and unlabeled data \citep{Chapelle2006semisupervisedlearning}, which is also referred to as prediction-powered inference \citep{Angelopoulos2021gentleintroduction,Ilker2024predictionpowered}. 

In many applications, such unlabeled covariates are easy to gather. For example, in the United States, we may aim to estimate the ATE for the effect of a new scholarship. Although we may know the covariates for an enormous number of students, we can assign treatment, scholarship, to only a limited number of them. In such cases, the unlabeled covariates contain information about the population over which the treatment effect is averaged, even though they do not contain treatment indicators or outcomes. 

We find that, under appropriate conditions, using unlabeled data allows us to construct an ATE estimator whose asymptotic variance, or equivalently, asymptotic MSE, is smaller than that of an estimator that ignores unlabeled data, as shown by \citet{Hahn1998ontherole}. To support this finding, we develop an asymptotic efficiency bound, a lower bound on the asymptotic variance, when using unlabeled covariates, propose ATE estimators, and show that the resulting asymptotic variances match the efficiency bound. In the methodological and theoretical arguments, we consider two practical scenarios, called the one-sample and two-sample scenarios. In the one-sample scenario, we interpret the unlabeled covariates as part of a dataset with missing variables, outcomes and the treatment indicator. In the two-sample scenario, we assume that labeled and unlabeled data are two independent datasets. The distinction is important because the two scenarios lead to different tangent spaces, different efficient influence functions, and different ways of using the unlabeled covariates.

Our efficient ATE estimators are developed based on the efficient influence function implied by the efficiency bound. We then extend the same logic to ATT, where the target distribution depends on the propensity score and therefore requires an additional treatment-law correction. This object is also called a Neyman orthogonal score in the debiased machine learning literature \citep{Chernozhukov2018doubledebiased}. The Neyman orthogonal scores include nuisance parameters, regression functions and a Riesz representer, which must be estimated before obtaining the ATE estimators. For the Riesz representer estimation, we employ generalized Riesz regression in \citet{Kato2025directdebiased,Kato2025directbias}, which generalizes the Riesz regression in \citet{Chernozhukov2021automaticdebiased}. The unified treatment in \citet{Kato2026aunified} further clarifies how Bregman divergence, loss-link choices, automatic regressor balancing, and automatic Neyman orthogonalization are connected. We extend this generalized Riesz regression perspective to the semi-supervised setting.

We list our contributions as follows:
\begin{itemize}
    \item We develop efficiency bounds for regular ATE estimators in the one-sample and two-sample scenarios, which also yield the corresponding Neyman orthogonal scores. In the two-sample scenario, the efficient influence functions are centered within each stratum.
    \item We construct asymptotically efficient estimators using the Neyman orthogonal scores and show that their asymptotic variances match the efficiency bounds. The two-sample target mixture parameter is treated as part of the estimand, not as a data-adaptive tuning parameter.
    \item We extend generalized Riesz regression for estimating nuisance parameters, including the Riesz representer and regression functions, while explicitly incorporating unlabeled covariates into the Riesz representer objective.
    \item We derive an explicit efficiency-gain identity showing that, in the same-population case, unlabeled covariates reduce the covariate averaging component but not the residual outcome-noise component.
    \item We extend the efficient-score construction to ATT. The ATT estimator is a debiased ratio estimator, and the corresponding score contains an additional treatment-law correction because the treated covariate distribution is unknown.
\end{itemize}

\paragraph{Scope of the contribution.} The role of generalized Riesz regression in this paper is different from its role in the general framework of \citet{Kato2026aunified}. That framework studies how to fit Riesz representers under Bregman divergences and how loss-link choices induce automatic regressor balancing and automatic Neyman orthogonalization. In contrast, the present paper fixes a semi-supervised observation scheme and derives the efficient scores, efficiency bounds, and attainable estimators under that scheme. The new content is therefore the interaction between unlabeled covariates, stratum-specific efficiency theory, and ATE or ATT targets. The generalized Riesz regression objectives are used as implementable nuisance-learning devices for the representers implied by these efficient scores.

\paragraph{Related work.} The related topics of this study include debiased machine learning, efficiency under the two-sample case (stratified sampling scheme), treatment effect estimation with missing values, density-ratio estimation, and semi-supervised learning.

In treatment effect estimation, we typically aim to attain the $\sqrt{n}$-rate with the smallest asymptotic variance, or equivalently, asymptotic MSE. We provide an efficiency bound, which is a lower bound on the asymptotic variance among regular estimators. As discussed in \citet{Uehara2020offpolicy}, when there are two independent datasets, we cannot apply the usual efficiency bounds developed for a single dataset. To derive efficiency bounds in such settings, existing studies employ the efficiency theory under the stratified sampling scheme \citep{Wooldridge2001asymptoticproperties}. Using this scheme, efficiency bounds have been proposed for various settings, including multiple log data, active learning, learning from positive and unlabeled data, and external-validity problems. This study also employs this technique to develop efficiency bounds.

The efficiency bounds are derived from the efficient influence functions. Certain efficient influence functions take forms that allow the removal of bias caused by the estimation errors of the nuisance parameters. Debiased machine learning is a framework for estimating treatment effects by utilizing such properties \citep{Chernozhukov2018doubledebiased}. We refer to efficient influence functions with these properties as Neyman orthogonal scores. \citet{Chernozhukov2022automaticdebiased} reframes this framework by characterizing Neyman orthogonal scores using the Riesz representer. \citet{Chernozhukov2021automaticdebiased} proposes Riesz regression, an end-to-end method for estimating the Riesz representer. \citet{Kato2025directbias} and \citet{Kato2025directdebiased} propose generalized Riesz regression by regarding the Riesz representer estimation problem as Bregman divergence minimization \citep{Sugiyama2011densityratio}. \citet{Kato2026aunified} provides a broader formulation that relates Bregman-Riesz fitting to automatic regressor balancing and automatic Neyman orthogonalization. The present paper uses this machinery for a different purpose. We keep the sampling scheme explicit and derive the efficiency bounds induced by semi-supervised covariates, rather than starting from a fixed single-sample Riesz functional. This is why the two-sample result requires separate labeled and unlabeled influence functions, and why the ATT result requires a debiased denominator.

This study generalizes treatment effect estimation under covariate shift \citep{Uehara2020offpolicy,Kato2024activeadaptive} and in the positive-unlabeled (PU) learning setup \citep{Kato2025puate}. PU learning is a classical problem, originally studied in \citet{Imbens1996efficientestimation}, and recently reframed by \citet{duPlessis2015convexformulation} as a modern statistical machine learning framework. Our sampling scheme arguments are significantly inspired by the works in this literature.

This study is also related to semi-supervised regression \citep{Azriel2022semisupervised,Kawakita2013semisupervisedlearning} and treatment effect estimation with missing values \citep{Heckman1974shadowprices,Robins1994estimationregression,Kennedy2020efficientnonparametric}. These studies clarify how auxiliary covariates or missingness mechanisms can improve estimation. Our focus differs because the target is the semiparametric efficiency bound for causal effects when the auxiliary observations contain only covariates.

A further distinction is that the unlabeled observations in this paper affect the target covariate distribution rather than providing additional outcomes or surrogate outcomes. This distinction is important for efficiency. The unlabeled sample can improve the estimation of the covariate averaging component, but it cannot directly reduce the conditional outcome-noise component.

\section{Problem Setting}
\label{sec:setup}
In this section, we formulate our problem setting. We define potential outcomes and observations separately by following the Neyman–Rubin causal model \citep{Neyman1923surapplications,Rubin1974estimatingcausal}. Then, we define the evaluation covariate density and the two sampling scenarios. .

\subsection{Potential Outcomes}
There is a binary treatment $d \in\{1, 0\}$. Let us define the corresponding potential outcome by $Y(d)$. 
Let $X\in \mathcal{X}\subset \mathbb{R}^k$ be a $k$-dimensional covariate, where $\mathcal{X}$ is the space. For each $d\in\{1, 0\}$, assume that the conditional distribution of $Y(d)$ given $X$ has its density, and let $r_{Y(d), 0}(y(d)\mid X)$ be the probability density function. 

\subsection{Average Treatment Effect}
This study focuses on the estimation of the average treatment effect, which is the expected value of $Y(1) - Y(0)$. We take the expectation over a distribution whose covariate probability density is given by
\[\kappa_0(x).\]
We call it the evaluation covariate density. This density function can differ from $p_0(x)$. We make the assumptions for $\kappa_0(x)$ in the following sections.

Under a given covariate density $\kappa_0(x)$, the ATE is defined as follows:
\begin{align*}
    \tau_0 &\coloneqq \bbE_{\kappa_0}\bigsqb{Y(1) - Y(0)} \coloneqq \int y r_{Y(1), 0}(y\mid x)\kappa_0(x) \rmd y \rmd x - \int y r_{Y(0), 0}(y\mid x)\kappa_0(x) \rmd y\rmd x,
\end{align*}
where $\bbE_{\kappa_0}[\cdot]$ denotes the expectation taken over the distribution whose covariate density is $\kappa_0(x)$. 

\subsection{Average Treatment Effect on the Treated}
In addition to ATE, we consider the average treatment effect on the treated. For a given evaluation covariate density $\kappa_0$, define
\[
\rho_{0, \kappa} \coloneqq \bbE_{\kappa_0}\sqb{e_0(1\mid X)},
\]
where $e_0(1\mid X)=P(D=1\mid X)$. The ATT target under the evaluation density $\kappa_0$ is
\[
\tau^{\text{ATT}}_{0, \kappa} \coloneqq \frac{\bbE_{\kappa_0}\sqb{e_0(1\mid X)\tau_0(X)}}{\rho_{0, \kappa}}.
\]
When $\kappa_0=p_0$, this target is the usual ATT in the one-sample population. When $\kappa_0=\kappa_{0, \beta}$ in the two-sample scenario, it is the ATT for the evaluation population determined by the mixture density. This parameter is not obtained by replacing the ATE density with the treated covariate density alone, because the treated covariate density itself depends on the unknown propensity score. This feature is the source of the treatment-law correction in Section~\ref{sec:att_extension}.

\subsection{Observation}
This section defines the sample, that is, observations of $X$, $D$, and $Y$. To define observations rigorously, we need to consider the censoring setting carefully.
To discuss data augmentation within the theory of semiparametric efficiency, we introduce two DGPs. The first DGP is the one-sample scenario, where there is only one dataset, and in this dataset, treatment indicators and outcomes are observed only for a subset of units. We also refer to this setting as the censoring setting. The second DGP is the two-sample scenario, where there are two independent datasets; one of the datasets contains data with covariates, treatment indicator, and outcomes, while the other only contains covariates. We also refer to this setting as the case-control setting or the stratified sampling scheme. We define these two DGPs below.

\paragraph{One-sample scenario.}
In the one-sample scenario, we observe a single dataset $\calD$, defined as follows: 
\begin{align*}
    \calD \coloneqq \cb{\bigp{X_i, O_i, \widetilde{D}_i, \widetilde{Y}_i}}^n_{i=1}\text{ with }\bigp{X_i, O_i, \widetilde{D}_i, \widetilde{Y}_i} \iid p_0(x, o, \widetilde{d}, \widetilde{y}).
\end{align*}
where $O_i \in \{1, 0\}$ is an observation indicator, $\widetilde{D}_i \in \{1, 0, \text{NA}\}$, and $\widetilde{Y}_i$ is the observable treatment indicator and outcome, defined as
\begin{align*}
    \widetilde{D}_i & \coloneqq \mathbbm{1}[O_i = 1] D_i + \mathbbm{1}[O_i = 0] \text{NA},\\
    \widetilde{Y}_i &\coloneqq \mathbbm{1}[O_i = 1] Y_i + \mathbbm{1}[O_i = 0] \text{NA},
\end{align*}
$D_i \in \{1, 0\}$ is a treatment indicator, and $Y_i$ is the outcome defined as 
\[Y_i \coloneqq \mathbbm{1}[D_i = 1] Y_i(1) + \mathbbm{1}[D_i = 0] Y_i(0).\]
Here, $\text{NA}$ denotes a missing value. Equivalently, we can write $\widetilde{Y}_i$ as
\[\widetilde{Y}_i =  \mathbbm{1}\bigsqb{O_i = 1, \widetilde{D}_i = 1} Y_i(1) + \mathbbm{1}\bigsqb{O_i = 1, \widetilde{D}_i = 0} Y_i(0) + \mathbbm{1}\bigsqb{O_i = 0} \text{NA}.\]
In this setting, we assume $p_0(x) = \kappa_0(x)$. 

Note that $\widetilde{D}$ and $\widetilde{Y}$ are observable, while $Y_i$ and $D_i$ are not observable when $O_i = 0$. 

\paragraph{Two-sample scenario}
\label{sec:case_control_dgp}
In the two-sample scenario, we observe two stratified datasets, $\calD_{\text{L}}$ and $\calD_{\text{U}}$: 
\begin{align*}
    &\calD_{\text{L}} \coloneqq \cb{\bigp{X_j, D_j, Y_j}}^m_{j=1}\text{ with }\bigp{X_j, D_j, Y_j} \iid p_0(x, d, y)\text{ and }\\
    &\calD_{\text{U}} \coloneqq \cb{Z_k}^l_{k=1}\text{ with }\bigp{Z_k} \iid q_0(x),
\end{align*}
where $m$ and $l$ are the sample sizes of each dataset, and $Y_j$ is the observed outcome defined as 
\[Y_j = \mathbbm{1}[D_j = 1]Y_j(1) + \mathbbm{1}[D_j = 0]Y_j(0),\] 
and $D_j \in \{1, 0\}$ is a treatment indicator.

\begin{figure}[t]
    \centering
    \includegraphics[width=0.8\linewidth]{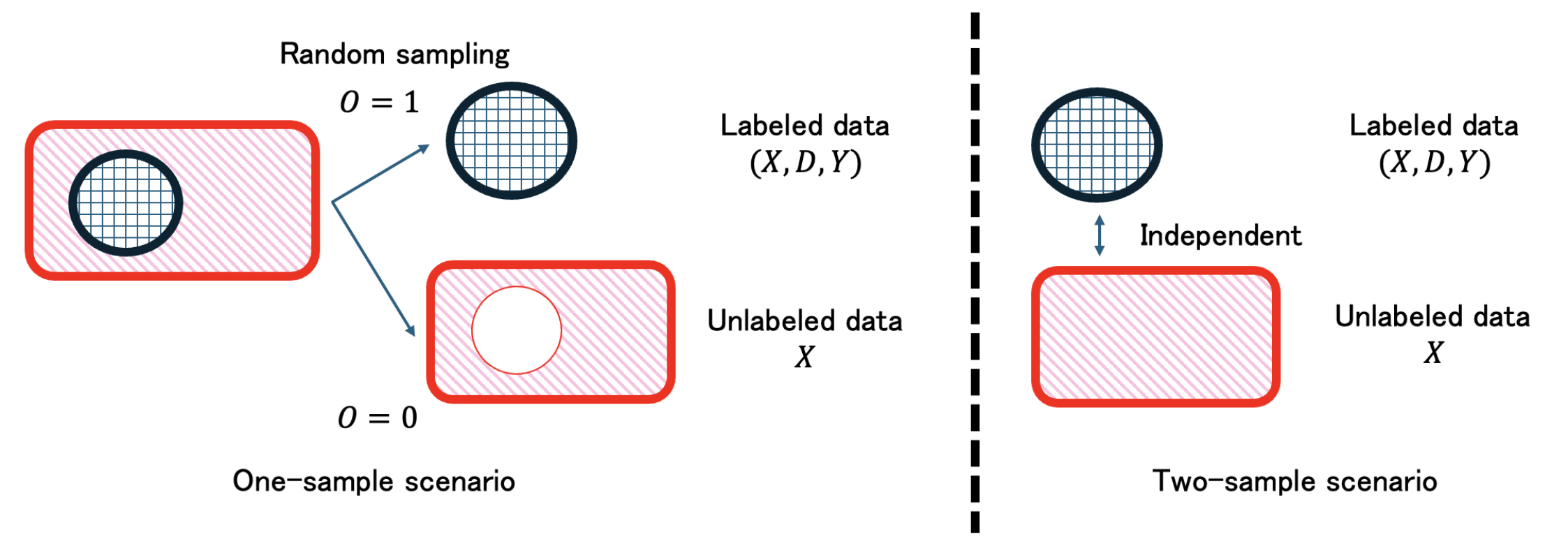}
    \caption{Illustration of the one-sample and two-sample scenarios.}
    \label{fig:osts_concept}
\end{figure}

\paragraph{Difference between the two settings}
We show an illustration that demonstrates the difference between the one-sample and two-sample scenarios in Figure~\ref{fig:osts_concept}. In both settings, we can identify and estimate the ATE in the standard way if we ignore the unlabeled auxiliary covariates. That is, in the one-sample scenario, we can estimate the ATE only by using $\calD$, while in the two-sample scenario, we can estimate the ATE only by using $\calD_{\text{L}}$. However, the unlabeled covariates change the information available about the evaluation covariate distribution. We demonstrate that this additional information reduces the asymptotic variance of efficient estimators.

A summary of the differences is provided below:
\begin{description}
    \item[One-sample scenario:] A single dataset is observed, where some observations do not include the treatment and outcome variables (i.e., contain only unlabeled covariates).
    \item[Two-sample scenario:] Two separate datasets are observed: one consists of fully labeled data, and the other contains only unlabeled covariates.
\end{description}

\begin{remark}[PU learning]
    Our terminology of the censoring and case-control settings comes from that in PU learning \citep{Niu2016theoreticalcomparisons}. In both settings, the goal is to learn a conditional class probability or a classifier using only positive and unlabeled data. In censoring PU learning, we consider one dataset from which labeled data are observed \citep{Elkan2008learningclassifiers}. In case-control PU learning, we assume that there exist two independent datasets, one labeled and the other unlabeled \citep{duPlessis2015convexformulation}. The case-control PU learning setting is also studied in \citet{Imbens1996efficientestimation}. 
\end{remark}

\subsubsection*{Notations and Assumptions} Throughout this study, let $P(R)$ denote the distribution of a random variable $R$. For simplicity, we assume that the distribution $P(R)$ of a continuous random variable $R$ has a probability density, whose notation depends on the random variable. For a probability density or mass function $p$, we denote the expectation over $p$ by $\bbE_p[\cdot]$. If the dependence is clear from the context, we omit $p$ and simply denote it as $\bbE[\cdot]$. Similarly, let $\text{Var}\p{\cdot}$ be the variance operator.
Let us denote the true mean and variance of $Y(d)$ conditioned on $X = x \in\mathcal{X}$ by $\mu_0(d, x) = \mathbb{E}[Y(d)\mid X=x]$ and
$\sigma^2_0(d, x) = \text{Var}(Y(d) \mid X=x)$, respectively. 

We make the following regularity assumption.
\begin{assumption}
\label{asm:dist}
There exist constants $\underline{C}$ and $\overline{C}$ such that $0 < \underline{C} < \overline{C} < \infty$ and for any $x\in\mathcal{X}$, $\big|\mu_0(d, x)\big| < \overline{C}$ and $\underline{C} < \sigma^2_0(d, x) < \overline{C}$ hold. 
\end{assumption}

\section{One-Sample Scenario}
\label{sec:onesample}
First, we consider the one-sample setting for the DGP, which is also referred to as the censoring setting. We redefine the DGP with its notations and assumptions in Section~\ref{sec:os_notationassumption}. Then, for this DGP, we develop an efficiency bound in Section~\ref{sec:os_efficiencybound}. We propose our estimator in Section~\ref{sec:os_ate_estimator} and show consistency in Section~\ref{sec:os_consistency} and asymptotic normality in Section~\ref{sec:os_asymp_prop}. 

\subsection{Notation and Assumption}
\label{sec:os_notationassumption}
This section introduces and summarizes the notations and assumptions, while recapping the DGP of the one-sample scenario. As defined in Section~\ref{sec:setup}, the DGP of this scenario is
\begin{align*}
    \calD \coloneqq \cb{\p{X_i, O_i, \widetilde{D}_i, \widetilde{Y}_i}}^n_{i=1}\text{ with }\p{X_i, O_i, \widetilde{D}_i, \widetilde{Y}_i} \in \calX \times \{1, 0\}\times \{1, 0, \text{NA}\} \times \cb{\calY \cup \text{NA}} \iid p_0(x, o, \widetilde{d}, \widetilde{y}).
\end{align*}
Let $\pi_0(o \mid X) = p(O = o\mid X)$ be the probability of observation $O = o$, $e_0(d\mid X) = P(D = d\mid X, O = 1)$ be the propensity score defined in the observed samples, and let $g_0(a\mid X) = P(O = 1, D = a\mid X) = e_0(a\mid X)\pi_0(1 \mid X)$ be the joint probability of $O = 1$ and $D = d$. Under this notation, the probability density $p_0(x, o, \widetilde{d}, \widetilde{y})$ is written as
\begin{align*}
    &p_0(x, o, \widetilde{d}, \widetilde{y}) = \\
    &p_0(x)\Bigp{\pi_0(0\mid X)}^{\mathbbm{1}[o = 0]} \Bigp{ g_0(1\mid x) r_{Y(1), 0}(\widetilde{y}\mid x)}^{\mathbbm{1}\sqb{o = 1, \widetilde{d} = 1}} \Bigp{g_0(0\mid x)r_{Y(0), 0}(\widetilde{y}\mid x)}^{\mathbbm{1}\sqb{o = 1, \widetilde{d} = 0}},
\end{align*}
For simplicity, we assume that the evaluation density $\kappa_0(x)$ is the marginal density of the covariates. 
\begin{assumption}[Evaluation density in the one-sample scenario]
\label{asm:os_eval_density}
The evaluation density $\kappa_0(x)$ is given as $\kappa_0(x) = p_0(x)$. 
\end{assumption}

We also make the following assumptions.
\begin{assumption}[Unconfoundedness and missing at random (MAR)]
\label{asm:os_unconfondedness}
    It holds that $(Y(1), Y(0))\indep D \mid X$ (unconfoundedness) and $(Y(1), Y(0)) \indep O \mid X$ (MAR). 
\end{assumption}
\begin{assumption}[Common support]
\label{asm:os_commonsupprt}
    There exists a universal constant $0 < \epsilon < 1/2$ such that for all $d\in\{1, 0\}$, $\epsilon < g_0(d\mid X) \leq 1 - \epsilon$ holds almost surely. 
\end{assumption}
Note that this assumption also implies the existence of a universal constant $0 < \epsilon' < 1/2$ such that $\epsilon' < \pi_0(1 \mid X)$.

\subsection{Efficiency Bound}
\label{sec:os_efficiencybound}
First, we derive the efficiency bound for regular estimators, which provides a lower bound on asymptotic variances. The efficiency bound is characterized via the efficient influence function \citep{Vaart1998asymptoticstatistics}. In this scenario, the influence function has the usual augmented inverse probability structure, with $g_0(d\mid X)$ replacing the propensity score because both treatment assignment and label observation must occur. The result is stated below, and the proof is provided in Appendix~\ref{appdx:proof:lem:os_efficiency_bound}.

\begin{lemma}
\label{lem:os_efficiency_bound}
    Suppose that Assumptions~\ref{asm:os_eval_density} through \ref{asm:os_commonsupprt} hold. Then, the efficient influence function is given as
    \[\psi^{\text{OS}}\bigp{X_i, O_i, \widetilde{D}_i, \widetilde{Y}_i; \mu_0, g_0, \tau_0},\] where 
    \begin{align*}
        \psi^{\text{OS}}\bigp{X_i, O_i, \widetilde{D}_i, \widetilde{Y}_i; \mu_0, g_0, \tau_0} &\coloneqq S^{\text{OS}}\bigp{X_i, O_i, \widetilde{D}_i, \widetilde{Y}_i; \mu_0, g_0} - \tau_0\\
        S^{\text{OS}}\bigp{X_i, O_i, \widetilde{D}_i, \widetilde{Y}_i; \mu_0, g_0} &\coloneqq \frac{\mathbbm{1}\bigsqb{O_i = 1, \widetilde{D}_i = 1}\bigp{\widetilde{Y}_i - \mu_0(1, X_i)}}{g_0(1\mid X)} - \frac{\mathbbm{1}\bigsqb{O_i = 1, \widetilde{D}_i = 0}\bigp{\widetilde{Y}_i - \mu_0(0, X_i)}}{g_0(0\mid X)}\\
        &+ \mu_0(1, X_i) - \mu_0(0, X_i).
    \end{align*}
\end{lemma}
Recall that $g_0(d\mid X)=\pi_0(1\mid X)e_0(d\mid X)$. The following proposition is Theorem~25.20 in \citet{Vaart1998asymptoticstatistics}, which connects the efficient influence function to the efficiency bound.

\begin{proposition}[Theorem~25.20 in \citet{Vaart1998asymptoticstatistics}.]
Let $R$ be some random variable, and $\calM$ be a model of its DGP. 
The (semiparametric) efficient influence function $\psi(R)$ is the gradient of $\theta$ with respect to the model $\mathcal{M}$, which has the smallest $L_2$-norm. It satisfies that for any regular estimator $\widehat{\theta}$ of a parameter of interest $\theta_0$ regarding a given parametric submodel, $\text{AMSE}\big(\widehat{\theta}\big)\geq \text{Var}\big(\psi(R)\big)$, where $\text{AMSE}\big(\widehat{\theta}\big)$ is the second moment of the limiting distribution of $\sqrt{n}\p{\widehat{\theta} - \theta_0}$. 
\end{proposition}

Using this proposition, we can derive the following efficiency bound from Lemma~\ref{lem:os_efficiency_bound}.

\begin{theorem}[Efficiency bound in the one-sample scenario]
If Assumptions~\ref{asm:os_eval_density} through \ref{asm:os_commonsupprt} hold, then the asymptotic variance of any regular estimator is lower bounded by 
\begin{align*}
    V^{\text{OS}} &\coloneqq \bbE\sqb{\psi^{\text{OS}}\p{X, O, \widetilde{D}, \widetilde{Y}; \mu_0, g_0, \tau_0}^2} = \bbE\sqb{\frac{\sigma^2_0(1, X)}{g_0(1\mid X)} + \frac{\sigma^2_0(0, X)}{g_0(0\mid X)} + \Bigp{\tau_0(X) - \tau_0}^2},
\end{align*}
where $\tau_0(X) \coloneqq \bbE\bigsqb{Y(1) - Y(0)\mid X}$ is the conditional ATE. 
\end{theorem}

Here, note that the efficient influence function depends on the unknown $\mu_0, g_0$, which are referred to as nuisance parameters. Since the efficient influence function satisfies the equation $\bbE\sqb{\psi^{\text{OS}}(X, O, Y; \mu_0, g_0, \tau_0)} = 0$, if the nuisance parameters are known and the exact expectation is computed, we can obtain $\tau_0$ by solving for $\tau_0$ that satisfies this equation. Thus, the efficient influence function provides a direct estimating equation for an efficient estimator. Furthermore, the accuracy of the estimation of the nuisance parameters affects the estimation of $\tau_0$, the parameter of interest.

\subsection{ATE Estimator}
\label{sec:os_ate_estimator}
Based on the efficient influence function, we propose an ATE estimator defined as 
\[\widehat{\tau}^{\text{OS}\mathchar`-\text{eff}}_n \coloneqq \frac{1}{n}\sum^n_{i=1}S^{\text{OS}}\p{X_i, O_i, \widetilde{D}_i, \widetilde{Y}_i; \widehat{\mu}_{n, i}, \widehat{g}_{n, i}},\] 
where $\widehat{\mu}_{n, i}$ and $\widehat{g}_{n, i}$ are estimators of $\mu_0$ and $g_0$. Note that the estimators can depend on $i$. 
This estimator is an extension of the augmented inverse probability weighting estimator, also called a doubly robust estimator \citep{Bang2005doublyrobust}. We say that an estimator is efficient if its asymptotic variance aligns with $V^{\text{OS}}$. 

For estimating the regression function $\mu_0$, we can employ methods for conditional ATE estimation \citep{Wager2018estimationinference,Curth2021nonparametricestimation,Kennedy2024minimaxrates}, as well as standard regression methods using parametric or nonparametric models \citep{Tsybakov2008introductionnonparametric,SchmidtHieber2020nonparametric}. We can also use targeted maximum likelihood estimation to refine this estimation \citep{vanderLaan2011targetedlearning}. 

For estimating $g_0$, we can use logistic regression or other advanced methods, such as the covariate balancing propensity score \citep{Imai2011estimationheterogeneous,Hainmueller2012entropybalancing} and Riesz regression \citep{Chernozhukov2021automaticdebiased}. As \citet{Zhao2019covariatebalancing}, \citet{BrunsSmith2025augmentedbalancing}, and \citet{Kato2025directbias} show, Riesz regression and covariate balancing methods are in a dual relationship, and Riesz regression can be interpreted as a special case of density ratio estimation \citep{Kato2025unifiedtheory}. For details, see Section~\ref{sec:os_ddml} and Appendix~\ref{appdx:os_ddml}.

\subsection{Generalized Riesz Regression}
\label{sec:os_ddml}
We explain how to construct estimators for $g_0$. In this study, we employ generalized Riesz regression, also referred to as Bregman-Riesz regression \citep{Kato2025directbias,Kato2025directdebiased,Kato2026aunified}. In the efficient estimation of causal parameters, Neyman orthogonal scores play an important role and typically correspond to the efficient score.
Specifically, asymptotically efficient estimators must be asymptotically linear with respect to the Neyman orthogonal scores. When the parameter of interest is linear in the regression functions, the Neyman orthogonal score can be decomposed into the Riesz representer and regression functions. In our semi-supervised setting, the Riesz representer is the component that determines how labeled residuals and unlabeled covariate averages are combined. In our framework, the Neyman orthogonal score is given by
\begin{align*}
    \psi^{\text{OS}}\bigp{X_i, O_i, \widetilde{D}_i, \widetilde{Y}_i; \mu_0, \alpha_0, \tau_0} &\coloneqq \alpha_0\p{O_i, \widetilde{D}_i, X_i}\p{\widetilde{Y}_i - \mu_0\p{\widetilde{D}_i, X_i}} + \mu_0(1, X_i) - \mu_0(0, X_i) - \tau_0,
\end{align*}
where we replace $g_0$ with $\alpha_0$ in the original definition of $\psi^{\text{OS}}$, and $\alpha_0\p{O_i, \widetilde{D}_i, X_i} \coloneqq \frac{\mathbbm{1}[O_i = 1, \widetilde{D}_i = 1]}{g_0(1\mid X_i)} - \frac{\mathbbm{1}[O_i = 1, \widetilde{D}_i = 0]}{g_0(0\mid X_i)}$ is the Riesz representer. Riesz regression, as proposed by \citet{Chernozhukov2021automaticdebiased}, is a method for estimating the Riesz representer in an end-to-end manner. \citet{Kato2025directbias} shows that Riesz regression is a specific instance of density ratio estimation and can be generalized via Bregman divergence minimization \citep{Sugiyama2011densityratio}. \citet{Kato2025directdebiased} further reformulates and extends this approach as direct debiased machine learning (DDML) via generalized Riesz regression. The efficiency results below use high-level product-rate assumptions for the learned nuisance functions. Generalized Riesz regression is one way to construct the representer estimator under these assumptions, while detailed rates for RKHS and neural network classes can be imported from the general theory of Bregman-Riesz fitting.

\paragraph{Generalized Riesz regression.} Generalized Riesz regression estimates $\alpha_0$ by minimizing the Bregman divergence between the true Riesz representer $\alpha_0$ and its model $\alpha$. That is, the estimation error of $\alpha_0$ is measured using the Bregman divergence. The recent unified framework of \citet{Kato2026aunified} emphasizes that the choice of Bregman divergence and link function determines both the geometry of the fitted representer and the associated balancing interpretation. For a twice differentiable convex function $f$ with bounded derivative, the population objective for Riesz representer estimation is written as
\begin{align}
    \label{eq:bdobj}
    &\text{BD}_{f}(\alpha)
    \coloneqq\\
    &\bbE\sqb{\mathbbm{1}[O = 1] \partial f\p{\alpha\p{O, \widetilde{D}, X}}\alpha\p{O, \widetilde{D}, X} - f\p{\alpha\p{O, \widetilde{D}, X}} - \Bigp{\partial f\p{\alpha\p{1, 1, X}} - \partial f\p{\alpha\p{1, 0, X}}}}.
\end{align}
The empirical counterpart $\widehat{\text{BD}}_{f}(\alpha)$ replaces expectations with sample averages. Here, we used $\tau_0 = \Exp{\Exp{\widetilde{Y}\mid O = 1, \widetilde{D} = 1, X} - \Exp{\widetilde{Y}\mid O = 1, \widetilde{D} = 0, X}}$. Minimizing this objective over a hypothesis class $\calA$ yields an estimator of $\alpha_0$, that is,
\[
\widehat{\alpha}^{\text{GRR}} \coloneqq \argmin_{\alpha \in \calA} \widehat{\text{BD}}_{f}(\alpha),
\]
where GRR denotes generalized Riesz regression.

The use of generalized Riesz regression allows us to naturally incorporate unlabeled covariates into the estimation of the Riesz representer. This is because, in \eqref{eq:bdobj}, 
we can approximate $\bbE\sqb{\Bigp{\partial f\p{\alpha\p{1, 1, X}} - \partial f\p{\alpha\p{1, 0, X}}}}$ using unlabeled covariates, whereas $\bbE\sqb{\partial f\p{\alpha\p{O, \widetilde{D}, X}}\alpha\p{O, \widetilde{D}, X}-f\p{\alpha\p{O, \widetilde{D}, X}}}$ requires labeled data. That is,
\begin{align*}
\widehat{\text{BD}}_{f}(\alpha)
&\coloneqq \frac{1}{n}\sum^n_{i=1}\mathbbm{1}[O_i = 1]\partial f\p{\alpha\p{O_i, \widetilde{D}_i, X_i}}\alpha\p{O_i, \widetilde{D}_i, X_i}-f\p{\alpha\p{O_i, \widetilde{D}_i, X_i}}\\
&- \frac{1}{n}\sum^n_{i=1}\Bigp{\partial f\p{\alpha\p{1, 1, X_i}} - \partial f\p{\alpha\p{1, 0, X_i}}},
\end{align*}
where the second term can be evaluated using both labeled and unlabeled data. This is the point at which the semi-supervised structure enters the Riesz representer estimation problem. Note that unlabeled covariates can be utilized even when $g_0$ is estimated via maximum likelihood. However, the generalized Riesz regression approach is arguably more appropriate in an end-to-end formulation because it directly targets the representer that appears in the Neyman orthogonal score. Also see \citet{Kawakita2013semisupervisedlearning}.

Let $\calA$ denote the model class for $\alpha_0$. If we set $f(\alpha) = (\alpha - 1)^2$, then
\begin{align*}
    &\text{BD}_{\text{LSIF}}\bigp{\alpha} \coloneqq \Exp{-2\bigp{\alpha\p{1, 1, X} - \alpha\p{1, 0, X}} + \mathbbm{1}[O = 1]\alpha\p{O, \widetilde{D}, X}^2}.
\end{align*}
This population objective corresponds to Riesz regression as in \citet{Chernozhukov2021automaticdebiased}. We refer to this objective as least squares Riesz (LS-Riesz) regression.

Now, redefine $\calA$ as the set of $\alpha$ such that $\alpha(1, 1, \cdot) > 1$ and $\alpha(1, 0, \cdot) < -1$, a condition that should hold under the common support assumption. For $f(\alpha) = (|\alpha| - 1)\log(|\alpha| - 1) + |\alpha|$ ($\alpha \in \calA$), the corresponding Bregman divergence is
\begin{align*}
    \text{BD}_{\text{UKL}}\bigp{\alpha} &\coloneqq \BigExp{\mathbbm{1}[O = 1]\p{\log\p{\abs{\alpha\p{O, \widetilde{D}, X}} - 1} + \abs{\alpha\p{O, \widetilde{D}, X}}}\\
    &- \log\bigp{\alpha(1, 1, X) - 1} - \log\bigp{-\alpha(1, 0, X) - 1}}.
\end{align*}
 We refer to this objective as Kullback-Leibler Riesz (KL-Riesz) regression, since the choice of $f$ yields the KL divergence.
 
By replacing the expectations with the sample mean and minimizing the empirical objective for $\alpha$, we can estimate $\alpha_0$. 

\paragraph{Interpretation.} As \citet{Kato2025directdebiased} discusses, LS-Riesz regression corresponds to the stable balancing weights proposed in \citet{Zubizarreta2015stableweights}, and KL-Riesz corresponds to the entropy balancing weights in \citet{Hainmueller2012entropybalancing}. These correspondences were originally shown in the covariate balancing literature, such as in \citet{Zhao2019covariatebalancing} and \citet{BrunsSmith2025augmentedbalancing}. They can be derived from duality relationships. \citet{Kato2026aunified} further interprets these relationships as automatic regressor balancing under suitable loss-link pairs.

Note that the duality depends on the model class used for $\alpha_0$, namely $\calA$. For the duality between LS-Riesz and stable balancing weights, linear models must be used for $\calA$, whereas for the duality between KL-Riesz and entropy balancing weights, logistic models for $\alpha_0$ are required. Therefore, the choice of $\calA$ is not a purely computational choice; it determines which balancing equations are targeted by the Riesz representer estimator.

\subsection{Consistency and double robustness}
\label{sec:os_consistency}
First, we prove the consistency result, that is, $\widehat{\tau}^{\text{OS}\mathchar`-\text{eff}}_n \xrightarrow{\rmp} \tau_0$ holds as $n\to \infty$. We can obtain this result relatively easily compared to asymptotic normality. We make the following assumption, which holds for most estimators of the nuisance parameters.

\begin{assumption}
\label{asm:consistency_censoring}
    There exist universal constants $\epsilon \in (0, 1/2), C \in (0, \infty)$ such that $\widehat{g}_{n, i}(a\mid X) \in (\epsilon, 1 - \epsilon)$ and $\widehat{\mu}_{n, i}(d\mid X) \in [-C, C]$ hold almost surely. 
    As $n \to \infty$, either of the following holds for all $i\in \{1,2,\dots,n\}$:
    \[\big\|\widehat{\mu}_{n, i} - \mu_0\big\|_2 = o_p(1)\text{ or }\big\|\widehat{g}_{n, i} - g_0\big\|_2 = o_p(1).\]
\end{assumption}

Then, the following consistency result holds. This result is given as a special case of Theorem~\ref{thm:os_asymp_normal}; therefore, we omit the proof. 
\begin{theorem}[Consistency in the one-sample setting]
    If Assumptions~\ref{asm:os_eval_density} through \ref{asm:os_commonsupprt}, and \ref{asm:consistency_censoring} hold, then $\widehat{\tau}^{\text{OS}\mathchar`-\text{eff}}_n \xrightarrow{\rmp} \tau_0$ holds as $n\to \infty$. 
\end{theorem}

This consistency structure is referred to as double robustness.

\subsection{Asymptotic Normality}
\label{sec:os_asymp_prop}
Next, we establish the asymptotic normality of our estimator. Unlike consistency, this requires stronger assumptions on the nuisance estimators, especially for the propensity score.

To prove asymptotic normality or $\sqrt{n}$-consistency, we must control the complexity of the nuisance parameter estimators. One simple approach is to assume the Donsker condition; however, it is well known that this condition often fails in high-dimensional regression settings. In such cases, asymptotic normality can still be attained using sample splitting, a common technique in this field \citep{Klaassen1987consistentestimation}, which has recently been refined by \citet{Chernozhukov2018doubledebiased} as cross-fitting.

\paragraph{Cross-fitting.}
We estimate $\mu_0$ and $g_0$ using cross-fitting. 
Cross-fitting is a variant of sample splitting \citep{Chernozhukov2018doubledebiased}. We randomly partition $\calD$ into $L > 0$ folds (subsamples), and for each fold $b \in \calL \coloneqq \{1,2,\dots, L\}$, the nuisance parameters are estimated using all other folds. Let the estimators for fold $b \in \calL$ be denoted by $\widehat{\mu}^{(b)}_n$ and $\widehat{g}^{(b)}_n$. Let $\calI^{(b)}$ be the index set of samples belonging to fold $b$. This construction separates nuisance estimation from score evaluation and avoids imposing a Donsker condition on the nuisance classes.

Various estimation methods may be used, including neural networks and Lasso, as long as they satisfy the convergence rate conditions in Assumption~\ref{asm:os_conv_rate}. The pseudocode is shown in Algorithm~\ref{alg:psudo_censoring}.

\begin{algorithm}[t]
\label{alg:psudo_censoring}
\caption{Cross-fitting in the one-sample scenario}
\begin{algorithmic}
\STATE Input: Observations $\calD \coloneqq \cb{\p{X_i, O_i, \widetilde{D}_i, \widetilde{Y}_i}}^n_{i=1}$, number of folds $L$, and estimation methods for $\mu_0$ and $g_0$. Let $\calI = \{1, 2, \dots, n\}$ be the index set. 
\STATE Randomly split $\calI$ into $L$ roughly equal-sized folds, $(\mathcal{I}^{(b)})_{b\in\calL}$. Note that $\bigcup_{b \in \calL}\calI^{(b)} = \calI$. 
\FOR{$b \in \calL$}
    \STATE Set the training data as $\mathcal{I}^{(-b)} = \{1,2,\dots,n\} \setminus \mathcal{I}^{(b)}$.
    \STATE Construct estimators of the nuisance parameters on $\mathcal{I}^{(-b)}$, denoted by $\widehat{\mu}^{(b)}_n$ and $\widehat{g}^{(b)}_n$.
\ENDFOR
\STATE Output: Obtain an ATE estimate $\widehat{\tau}^{\text{OS}\mathchar`-\text{eff}}_n$ using $\widehat{\mu}^{(b)}_n$ and $\widehat{g}^{(b)}_n$.
\end{algorithmic}
\end{algorithm}

\paragraph{Asymptotic normality.}
We present results for the case with cross-fitting, but similar results hold under the Donsker condition.

We make the following assumptions:
\begin{assumption}
\label{asm:os_conv_rate}
For each $b \in \calL$, as $n\to\infty$, the following hold:
    \begin{itemize}
        \item $\big\| \mu_0(a, X) - \widehat{\mu}^{(b)}_{n, i}(a, X) \big\|_2 = o_p(1)$ and $\big\| g_0(a\mid X) - \widehat{g}^{(b)}_{n, i}(a\mid X) \big\|_2 = o_p(1)$. 
        \item $\big\|\mu_0(a, X) - \widehat{\mu}^{(b)}_n(a, X)\big\|_2 \cdot \big\| g_0(a\mid X) - \widehat{g}^{(b)}_{n, i}(a\mid X) \big\|_2 = o_p(n^{-1/2})$ for $a \in \{1, 0\}$.
    \end{itemize}
\end{assumption}

We define the estimator as 
\[
\widehat{\tau}^{\text{OS}\mathchar`-\text{eff}}_n \coloneqq \frac{1}{n}\sum_{b \in \calL}\sum_{i \in \calI^{(b)}}S^{\text{OS}}\p{X_i, O_i, \widetilde{D}_i, \widetilde{Y}_i; \widehat{\mu}^{(b)}_n, \widehat{g}^{(b)}_n}
\]
and show that the asymptotic normality holds as follows:
\begin{theorem}[Asymptotic normality in the one-sample scenario]
    \label{thm:os_asymp_normal}
    Consider the one-sample scenario. Suppose Assumptions~\ref{asm:os_eval_density} through \ref{asm:os_commonsupprt} and \ref{asm:os_conv_rate} hold; that is, $\widehat{\mu}_{n, i} = \widehat{\mu}^{(b)}_n$ and $\widehat{g}_{n, i} = \widehat{g}^{(b)}_n$ are constructed via cross-fitting with suitable convergence rates. Then,
    \begin{align*}
        \sqrt{n}\p{\widehat{\tau}^{\text{OS}\mathchar`-\text{eff}}_n - \tau_0} \xrightarrow{\rmd} \mathcal{N}(0, V^{\text{OS}})\text{ as }n\to \infty. 
    \end{align*}
\end{theorem}
The proof is provided in Appendix~\ref{appdx:normal_cens}. The asymptotic variance of $\widehat{\tau}^{\text{OS}\mathchar`-\text{eff}}_n$ matches the efficiency bound. Therefore, Theorem~\ref{thm:os_asymp_normal} also implies that $\widehat{\tau}^{\text{OS}\mathchar`-\text{eff}}_n$ is asymptotically efficient.

For inference, let $\widehat{\psi}^{\text{OS}}_i$ denote the influence function in Lemma~\ref{lem:os_efficiency_bound} evaluated at the cross-fitted nuisance estimators and at $\widehat{\tau}^{\text{OS}\mathchar`-\text{eff}}_n$. We estimate the asymptotic variance by
\[
\widehat{V}^{\text{OS}} \coloneqq \frac{1}{n}\sum^n_{i=1}\Bigp{\widehat{\psi}^{\text{OS}}_i - \frac{1}{n}\sum^n_{i'=1}\widehat{\psi}^{\text{OS}}_{i'}}^2.
\]
The resulting Wald interval uses $\widehat{V}^{\text{OS}}/n$ as the variance of $\widehat{\tau}^{\text{OS}\mathchar`-\text{eff}}_n$.

We now discuss alternative ATE estimators.
\begin{remark}[Inefficiency of the Inverse Probability Weighting (IPW) estimator]
\label{rem:IPW}
    The IPW estimator is defined as 
    \[
    \widehat{\tau}^{\text{OS}\mathchar`-\text{IPW}}_n \coloneqq \frac{1}{n}\sum^n_{i=1}\p{\frac{\mathbbm{1}[O_i = 1, \widetilde{D}_i = 1]\widetilde{Y}_i}{\widehat{g}_{n, i}(1\mid X_i)} - \frac{\mathbbm{1}[O_i = 1, \widetilde{D}_i = 0]\widetilde{Y}_i}{\widehat{g}_{n, i}(0\mid X_i)}}.
    \] 
    Unlike our proposed efficient estimator, this estimator does not use the conditional outcome estimators \citep{Horvitz1952generalization}. When $g_0$ and $\pi_0$ are known, it is unbiased. However, it suffers from a large asymptotic variance:
    \[
    V^{\text{IPW}} \coloneqq \bbE\sqb{\frac{\bbE[Y(1)^2\mid X]}{g_0(1\mid X)}+ \frac{\bbE[Y(0)^2\mid X]}{g_0(0\mid X)}}.
    \]
    Here, $V^{\text{IPW}} \geq V^{\text{OS}}$, with equality when $\mu_0(d, x)$ is zero for all $x$. Thus, the IPW estimator is inefficient relative to $\widehat{\tau}^{\text{OS}\mathchar`-\text{eff}}_n$. Moreover, if $\pi_0$ is unknown, stronger assumptions are needed to establish asymptotic normality compared to our efficient estimator.
\end{remark}

\begin{remark}[Regression Adjustment (RA) estimator]
\label{rem:DM}
Another alternative is the RA estimator, defined as 
\[
\widehat{\tau}^{\text{OS}\mathchar`-\text{RA}}_n \coloneqq \frac{1}{n}\sum^n_{i=1}\widehat{\mu}_{n, i}(1, X_i) - \widehat{\mu}_{n, i}(0, X_i),
\]
also known as the naive plug-in or direct method estimator. Its asymptotic normality heavily depends on the estimators $\widehat{\mu}_{n, i}$. 
\end{remark}

\section{Two-Sample Scenario}
Next, we consider the two-sample scenario for the DGP, which is also referred to as the case-control setting and stratified sampling scheme. We reintroduce the notation and assumptions required for our analysis in Section~\ref{sec:ts_notationassumption}. Section~\ref{sec:ts_efficiencybound} presents the efficiency bound, and Section~\ref{sec:ts_ate_estimator} provides an ATE estimator under this setting. We establish consistency in Section~\ref{sec:ts_consistency} and asymptotic normality in Section~\ref{sec:ts_asymp_norm}. Finally, we compare the one-sample and two-sample scenarios in Section~\ref{sec:osts_comp}.

\subsection{Notation and Assumptions}
\label{sec:ts_notationassumption}
As introduced in Section~\ref{sec:setup}, the DGP for the two-sample scenario is defined as
\begin{align*}
    &\calD_{\text{L}} \coloneqq \cb{\bigp{X_j, D_j, Y_j}}^m_{j=1}, \text{ with } \bigp{X_j, D_j, Y_j} \in \calX \times \{1, 0\} \times \calY, \text{ and } \bigp{X_j, D_j, Y_j} \iid p_0(x, d, y),\\
    &\calD_{\text{U}} \coloneqq \cb{Z_k}^l_{k=1}, \text{ with } Z_k \in \calX, \text{ and } Z_k \iid q_0(x).
\end{align*}
Let $e_0(d\mid X) = P(D = d\mid X)$ denote the propensity score. Then, the joint density $p_0(x, d, y)$ can be written as
\begin{align*}
    p_0(x, d, y) = p_0(x)\Bigp{e_0(1\mid x) r_{Y(1), 0}(y\mid x)}^{\mathbbm{1}[d = 1]} \Bigp{e_0(0\mid x)r_{Y(0), 0}(y\mid x)}^{\mathbbm{1}[d = 0]}.
\end{align*}

For the evaluation density, we make the following assumption. 
\begin{assumption}[Evaluation density in the two-sample scenario]
\label{asm:ts_eval_density}
There exists a fixed $\beta \in [0, 1]$ such that
    \[\kappa_0(x) = \kappa_{0, \beta}(x) = \beta p_0(x) + (1 - \beta) q_0(x).\]
The value of $\beta$ is part of the estimand and is chosen by the researcher. It is not selected by minimizing an asymptotic variance unless the target density is invariant to $\beta$.
\end{assumption}

The following support condition is needed because all outcome and treatment information is contained in the labeled sample.
\begin{assumption}[Support]
\label{asm:ts_support}
It holds that $\kappa_{0, \beta} \ll p_0$. In addition, there exists a universal constant $C < \infty$ such that $\kappa_{0, \beta}(X)/p_0(X) \leq C$ almost surely under $p_0$.
\end{assumption}
If $\beta < 1$, Assumption~\ref{asm:ts_support} implies $q_0 \ll p_0$. Without this condition, there may be target covariate regions for which outcomes are never observed in the labeled sample.

We also impose the following assumptions.
\begin{assumption}[Unconfoundedness]
\label{asm:ts_unconfondedness}
    The potential outcomes satisfy $(Y(1), Y(0))\indep D \mid X$. 
\end{assumption}
\begin{assumption}[Common support]
\label{asm:ts_commonsupprt}
    There exists a universal constant $0 < \epsilon < 1/2$ such that $\epsilon  < e_0(1\mid X) < 1 - \epsilon$ almost surely under $p_0$. 
\end{assumption}

Define
\[
\omega_{0, \beta}(x) \coloneqq \frac{\kappa_{0, \beta}(x)}{p_0(x)},
\]
and
\[
v_{0, \beta}(d, x) \coloneqq \frac{e_0(d\mid x)}{\omega_{0, \beta}(x)} = \frac{p_0(d, x)}{\kappa_{0, \beta}(x)}.
\]
Thus, $1/v_{0, \beta}(d, x)$ is the efficient residual weight for the labeled sample. We also define
\[
\tau_{p,0} \coloneqq \bbE_{p_0}\sqb{\tau_0(X)}, \tau_{q,0} \coloneqq \bbE_{q_0}\sqb{\tau_0(Z)}, \tau_0 = \beta \tau_{p,0} + (1 - \beta)\tau_{q,0}.
\]

\subsection{Efficiency Bound}
\label{sec:ts_efficiencybound}
Following \citet{Uehara2020offpolicy}, we derive the efficiency bound using the efficiency arguments under the two-sample scenario. In this scheme, there are two efficient influence functions, one for the labeled stratum and one for the unlabeled stratum. The proof is provided in Appendix~\ref{appdx:proof:lem:ts_efficiency_bound}. 

\begin{lemma}
\label{lem:ts_efficiency_bound}
If Assumptions~\ref{asm:ts_eval_density} through \ref{asm:ts_commonsupprt} hold, then the efficient influence functions for the labeled and unlabeled strata are given by
\begin{align*}
    \psi^{\text{TS}}_{\text{L}}(X, D, Y; \mu_0, v_{0, \beta}) &\coloneqq S^{\text{TS}}_{(X, D, Y)}\p{X, D, Y; \mu_0, v_{0, \beta}} + \beta\p{\tau_0(X) - \tau_{p,0}},\\
    \psi^{\text{TS}}_{\text{U}}(Z; \mu_0) &\coloneqq (1 - \beta)\p{\tau_0(Z) - \tau_{q,0}},\\
    S^{\text{TS}}_{(X, D, Y)}(X, D, Y; \mu_0, v_{0, \beta}) &\coloneqq \frac{\mathbbm{1}\bigsqb{D = 1}\bigp{Y - \mu_0(1, X)}}{v_{0, \beta}(1, X)} - \frac{\mathbbm{1}\bigsqb{D = 0}\bigp{Y - \mu_0(0, X)}}{v_{0, \beta}(0, X)}.
\end{align*}
\end{lemma}
The centering in Lemma~\ref{lem:ts_efficiency_bound} is stratum specific. The labeled stratum is centered by $\tau_{p,0}$ and the unlabeled stratum is centered by $\tau_{q,0}$. This is necessary because each stratum has its own sampling distribution.

As in the one-sample scenario, the efficient influence functions directly yield the following efficiency bound. 

\begin{theorem}[Efficiency bound in the two-sample scenario]
\label{thm:efficy_bound_cc}
Let $N = m + l$, where $m = \alpha N$ and $l = (1 - \alpha)N$ for some $\alpha \in (0, 1)$. If Assumptions~\ref{asm:ts_eval_density} through \ref{asm:ts_commonsupprt} hold, then the asymptotic variance of any regular estimator is lower bounded by
\begin{align*}
V^{\text{TS}}(\beta)
&\coloneqq \frac{1}{\alpha}\bbE_{p_0}\sqb{\psi^{\text{TS}}_{\text{L}}(X, D, Y; \mu_0, v_{0, \beta})^2} + \frac{1}{1 - \alpha}\bbE_{q_0}\sqb{\psi^{\text{TS}}_{\text{U}}(Z; \mu_0)^2}\\
&= \frac{1}{\alpha}\bbE_{p_0}\sqb{\p{\frac{\sigma^2_0(1, X)}{e_0(1\mid X)} + \frac{\sigma^2_0(0, X)}{e_0(0\mid X)}}\p{\frac{\kappa_{0, \beta}(X)}{p_0(X)}}^2}\\
&{}+ \frac{\beta^2}{\alpha}\bbE_{p_0}\sqb{\Bigp{\tau_0(X) - \tau_{p,0}}^2} + \frac{(1 - \beta)^2}{1 - \alpha}\bbE_{q_0}\sqb{\Bigp{\tau_0(Z) - \tau_{q,0}}^2}.
\end{align*}
\end{theorem}

The parameter $\beta$ is fixed throughout the theorem. If $p_0 = q_0$, all values of $\beta$ define the same evaluation density. Only in such cases can $\beta$ be selected to improve precision without changing the estimand.

\subsection{ATE Estimator}
\label{sec:ts_ate_estimator}
Based on the efficient influence functions, we define the estimator as
\[
\widehat{\tau}^{\text{TS}\mathchar`-\text{eff}}_n \coloneqq \frac{1}{m}\sum^m_{j=1} S^{\text{TS}}_{(X, D, Y)}\bigp{X_j, D_j, Y_j; \widehat{\mu}, \widehat{v}_{\beta}} + \beta\frac{1}{m}\sum^m_{j=1} S^{\text{TS}}_{(X)}\bigp{X_j; \widehat{\mu}} + \p{1 - \beta}\frac{1}{l}\sum^l_{k=1} S^{\text{TS}}_{(X)}\bigp{Z_k; \widehat{\mu}},
\]
where $S^{\text{TS}}_{(X)}(x; \mu) \coloneqq \mu(1, x) - \mu(0, x)$. Here, $\widehat{\mu}$ and $\widehat{v}_{\beta}$ denote estimators of $\mu_0$ and $v_{0, \beta}$. Unlike in the one-sample scenario, we do not use the observation indicator $O$, since it is deterministically known whether a unit belongs to the labeled or unlabeled stratum. This distinction leads to theoretical differences from the one-sample scenario.

\subsection{Generalized Riesz Regression in the Two-Sample Scenario}
\label{sec:ts_grr}
The two-sample efficient score also yields a semi-supervised generalized Riesz regression objective. The Riesz representer for the labeled residual part is
\[
\alpha_{0, \beta}(D, X) \coloneqq \frac{\mathbbm{1}\bigsqb{D = 1}}{v_{0, \beta}(1, X)} - \frac{\mathbbm{1}\bigsqb{D = 0}}{v_{0, \beta}(0, X)}.
\]
It represents the linear functional $h \mapsto \bbE_{\kappa_{0, \beta}}\sqb{h(1, X) - h(0, X)}$ through expectations under the labeled distribution. For a twice differentiable convex function $f$, define
\begin{align*}
\text{BD}^{\text{TS}}_{f, \beta}(\alpha)
&\coloneqq \bbE_{p_0}\sqb{\partial f\p{\alpha(D, X)}\alpha(D, X) - f\p{\alpha(D, X)}}\\
&{}- \beta\bbE_{p_0}\sqb{\partial f\p{\alpha(1, X)} - \partial f\p{\alpha(0, X)}}\\
&{}- (1 - \beta)\bbE_{q_0}\sqb{\partial f\p{\alpha(1, Z)} - \partial f\p{\alpha(0, Z)}}.
\end{align*}
The empirical counterpart is
\begin{align*}
\widehat{\text{BD}}^{\text{TS}}_{f, \beta}(\alpha)
&\coloneqq \frac{1}{m}\sum^m_{j=1}\Bigp{\partial f\p{\alpha(D_j, X_j)}\alpha(D_j, X_j) - f\p{\alpha(D_j, X_j)}}\\
&{}- \beta\frac{1}{m}\sum^m_{j=1}\Bigp{\partial f\p{\alpha(1, X_j)} - \partial f\p{\alpha(0, X_j)}}\\
&{}- (1 - \beta)\frac{1}{l}\sum^l_{k=1}\Bigp{\partial f\p{\alpha(1, Z_k)} - \partial f\p{\alpha(0, Z_k)}}.
\end{align*}
Then, we estimate $\alpha_{0, \beta}$ by
\[
\widehat{\alpha}^{\text{TS}\mathchar`-\text{GRR}} \in \argmin_{\alpha\in\calA}\cb{\widehat{\text{BD}}^{\text{TS}}_{f, \beta}(\alpha) + \lambda J(\alpha)}.
\]
This objective makes explicit how the unlabeled covariates enter generalized Riesz regression: the labeled sample controls the residual representer geometry, while both labeled and unlabeled covariates define the target linear functional. This connection follows the generalized Riesz regression perspective, where Bregman divergence and loss-link choices determine both representer fitting and automatic balancing behavior \citep{Kato2026aunified}.

\subsection{Consistency}
\label{sec:ts_consistency}
We impose the following assumption.
\begin{assumption}
\label{asm:ts_consistency}
As $m, l \to \infty$, it holds that $\big\|\widehat{\mu} - \mu_0\big\|_2 = o_p(1)$ or $\big\|\widehat{v}_{\beta} - v_{0, \beta}\big\|_2 = o_p(1)$.
\end{assumption}

Then, the following consistency result holds.
\begin{theorem}[Consistency in the two-sample scenario]
If Assumptions~\ref{asm:ts_eval_density} through \ref{asm:ts_consistency} hold, then $\widehat{\tau}^{\text{TS}\mathchar`-\text{eff}}_n \xrightarrow{\rmp} \tau_0$ as $N \to \infty$.
\end{theorem}

\subsection{Asymptotic Normality}
\label{sec:ts_asymp_norm}
Next, we establish the asymptotic normality of the estimator. 
\begin{assumption}
\label{asm:ts_conv_rate}
For each $b \in \calL$, as $m, l \to \infty$, the following hold: for each $d\in\{1, 0\}$, 
\begin{align*}
    &\big\| \mu_0(d, X) - \widehat{\mu}^{(b)}(d, X) \big\|_2 = o_p(1), \big\| v_{0, \beta}(d\mid X) - \widehat{v}^{(b)}_{\beta}(d\mid X) \big\|_2 = o_p(1),\\
    &\big\| \mu_0(d, X) - \widehat{\mu}^{(b)}(d, X) \big\|_2\big\| v_{0, \beta}(d\mid X) - \widehat{v}^{(b)}_{\beta}(d\mid X) \big\|_2 = o_p(1/\sqrt{\min\{m, l\}}).
\end{align*}
\end{assumption}

We now establish asymptotic normality in the following theorem, with the proof provided in Appendix~\ref{appdx:normal_cc}. In this result, we consider the asymptotic regime where the sample sizes $m$ and $l$ approach infinity while maintaining a fixed ratio $m:l = \alpha:(1-\alpha)$. 

\begin{theorem}[Asymptotic normality in the two-sample scenario]
\label{thm:ts_asymp_norm}
Let $N = m + l$. Fix $\alpha \in (0, 1)$. Consider the two-sample scenario with sample sizes $m, l$ such that $m = \alpha N$ and $l = (1-\alpha)N$. Suppose that Assumptions~\ref{asm:ts_eval_density} through \ref{asm:ts_consistency} and \ref{asm:ts_conv_rate} hold. Also assume that nuisance estimators are constructed via cross-fitting. Then,
\[
\sqrt{N}\p{\widehat{\tau}^{\text{TS}\mathchar`-\text{eff}}_n - \tau_0} \xrightarrow{\rmd} \mathcal{N}\p{0, V^{\text{TS}}(\beta)}\text{ as }N\to \infty.
\] 
\end{theorem}

Thus, the proposed estimator is efficient with respect to the efficiency bound derived in Theorem~\ref{thm:efficy_bound_cc}.

For inference, let $\widehat{\psi}^{\text{TS}}_{\text{L},j}$ and $\widehat{\psi}^{\text{TS}}_{\text{U},k}$ denote the two stratum influence functions in Lemma~\ref{lem:ts_efficiency_bound} evaluated at the cross-fitted nuisance estimators and at $\widehat{\tau}^{\text{TS}\mathchar`-\text{eff}}_n$. We estimate the scaled variance by
\begin{align*}
\widehat{V}^{\text{TS}}(\beta)
&\coloneqq \frac{N}{m}\frac{1}{m}\sum^m_{j=1}\Bigp{\widehat{\psi}^{\text{TS}}_{\text{L},j}-\frac{1}{m}\sum^m_{j'=1}\widehat{\psi}^{\text{TS}}_{\text{L},j'}}^2\\
&{}+\frac{N}{l}\frac{1}{l}\sum^l_{k=1}\Bigp{\widehat{\psi}^{\text{TS}}_{\text{U},k}-\frac{1}{l}\sum^l_{k'=1}\widehat{\psi}^{\text{TS}}_{\text{U},k'}}^2.
\end{align*}
The resulting Wald interval uses $\widehat{V}^{\text{TS}}(\beta)/N$ as the variance of $\widehat{\tau}^{\text{TS}\mathchar`-\text{eff}}_n$.

\begin{corollary}
\label{cor:same_population_ts}
If $p_0(X) = q_0(X)$ almost surely, then every $\beta\in[0,1]$ defines the same evaluation density. In this case, $V^{\text{TS}}(\beta)$ is minimized at $\beta = \alpha$, and
\[
V^{\text{TS}}(\alpha) = \frac{1}{\alpha}\bbE_{p_0}\sqb{\frac{\sigma^2_0(1, X)}{e_0(1\mid X)} + \frac{\sigma^2_0(0, X)}{e_0(0\mid X)}} + \bbE_{p_0}\sqb{\Bigp{\tau_0(X) - \tau_0}^2}.
\]
\end{corollary}

\subsection{Comparison with the One-Sample Scenario}
\label{sec:osts_comp}
The difference between the one-sample and two-sample scenarios appears in the formulation of ATE estimators, the setup of Riesz regression, and the corresponding efficiency arguments. In the one-sample scenario, the observation indicator is random within a single superpopulation, and the efficient influence function uses $g_0(d\mid X) = P(O = 1, D = d\mid X)$. In the two-sample scenario, the sample membership is fixed by design, and the efficient influence function has separate labeled and unlabeled stratum components. This distinction is the reason why the two-sample bound in Theorem~\ref{thm:efficy_bound_cc} uses stratum-specific centering.

\section{Many Unlabeled Data}
\label{sec:infintiteauxilarylower}
In many applications, we often have access to many more unlabeled data points than fully labeled ones, as unlabeled covariates are less costly to collect. The main results above use the fixed-ratio regime $m/N\to\alpha$. This section records the corresponding sequential implication when the unlabeled sample is asymptotically much larger than the labeled sample.

\subsection{Two-Sample Scenario}
In the two-sample scenario, let $l/m\to\infty$ and normalize by $\sqrt{m}$. Then, the unlabeled stratum average is negligible, but the labeled stratum still contains both the residual component and the $p_0$-covariate averaging component when $\beta>0$.

\begin{corollary}
\label{lem:ts_asymp_norminfinitemany}
Assume the same conditions as in Theorem~\ref{thm:ts_asymp_norm}. Let $l/m\to\infty$. Then,
\begin{align*}
    \sqrt{m}\p{\widehat{\tau}^{\text{TS}\mathchar`-\text{eff}}_n - \tau_0} \xrightarrow{\rmd} \mathcal{N}\p{0, \widetilde{V}^{\text{TS}}(\beta)},
\end{align*}
where
\[
\widetilde{V}^{\text{TS}}(\beta) \coloneqq \bbE_{p_0}\sqb{\p{\frac{\sigma^2_0(1, X)}{e_0(1\mid X)} + \frac{\sigma^2_0(0, X)}{e_0(0\mid X)}}\p{\frac{\kappa_{0, \beta}(X)}{p_0(X)}}^2} + \beta^2\bbE_{p_0}\sqb{\Bigp{\tau_0(X) - \tau_{p,0}}^2}.
\]
In the target-domain case $\beta = 0$, only the residual component remains under this normalization.
\end{corollary}

\subsection{One-Sample Scenario}
The one-sample analogue requires a triangular-array formulation if the probability of observing labels changes with the sample size. To avoid conflating this regime with the fixed-DGP efficiency theory in Section~\ref{sec:setup}, we state the qualitative implication only. If an external covariate sample makes the empirical distribution of $X$ negligible relative to the labeled sample size, then the covariate averaging component in the one-sample efficiency bound is estimated with negligible additional noise, while the conditional outcome-noise component remains governed by the labeled observations. A fully formal triangular-array statement can be added separately if this regime is the focus of the application.

\section{Efficiency Gain}
By using auxiliary unlabeled covariates, we can reduce the asymptotic variance of ATE estimators. As \citet{Hahn1998ontherole} shows, for a labeled dataset $\cb{\bigp{X_i, D_i, Y_i}}^{n^\dagger}_{i=1}$, the efficiency bound of ATE estimators $\widehat{\tau}$ is given as $V^\dagger \coloneqq \bbE\sqb{\frac{\sigma^2_0(1, X)}{P(D = 1\mid X)} + \frac{\sigma^2_0(0, X)}{P(D = 0\mid X)}} + \bbE\sqb{\bigp{\tau_0(X) - \tau_0}^2}$, and an efficient ATE estimator satisfies
\[
\sqrt{n^\dagger}\p{\widehat{\tau} - \tau_0}\xrightarrow{\rmd} \calN\p{0, V^\dagger}.
\]
The corrected two-sample bound makes the efficiency gain transparent in the same-population case $p_0 = q_0$. If $m/N\to\alpha$ and $\beta = \alpha$, then the semi-supervised bound is
\[
V^{\text{TS}}(\alpha) = \frac{1}{\alpha}\bbE_{p_0}\sqb{\frac{\sigma^2_0(1, X)}{e_0(1\mid X)} + \frac{\sigma^2_0(0, X)}{e_0(0\mid X)}} + \bbE_{p_0}\sqb{\Bigp{\tau_0(X) - \tau_0}^2}.
\]
By contrast, using only the $m$ labeled observations and scaling by $\sqrt{N}$ gives
\[
V^{\text{sup}} = \frac{1}{\alpha}\bbE_{p_0}\sqb{\frac{\sigma^2_0(1, X)}{e_0(1\mid X)} + \frac{\sigma^2_0(0, X)}{e_0(0\mid X)} + \Bigp{\tau_0(X) - \tau_0}^2}.
\]
Therefore,
\[
V^{\text{sup}} - V^{\text{TS}}(\alpha) = \p{\frac{1}{\alpha} - 1}\bbE_{p_0}\sqb{\Bigp{\tau_0(X) - \tau_0}^2}.
\]
This identity is the main variance-reduction message. Unlabeled covariates reduce the covariate averaging component but do not reduce the residual outcome-noise component.

The same message holds for ATT. In the same-population case $p_0=q_0$ with $\beta=\alpha$, let $\rho_0=\bbE_{p_0}\sqb{e_0(X)}$ and $\Delta_0(X)=\tau_0(X)-\tau^{\text{ATT}}_0$. The labeled-only ATT bound under $\sqrt{N}$ scaling contains $\frac{1}{\alpha\rho_0^2}\bbE_{p_0}\sqb{e_0(X)^2\Delta_0(X)^2}$ as the treated-covariate averaging component. The semi-supervised ATT bound contains the same component without the factor $1/\alpha$. Therefore,
\[
V^{\text{ATT}\mathchar`-\text{sup}} - V^{\text{ATT}\mathchar`-\text{TS}}(\alpha) = \p{\frac{1}{\alpha}-1}\frac{\bbE_{p_0}\sqb{e_0(X)^2\Delta_0(X)^2}}{\rho_0^2}.
\]
Thus, for ATT, unlabeled covariates reduce the treated-covariate averaging component weighted by the squared propensity score.

\section{Covariate Shift Adaptation}
\citet{Uehara2020offpolicy} investigates ATE estimation, equivalently, off-policy evaluation, under covariate shift. Our formulation includes their ATE estimation approach as a special case. When $\beta = 0$ in the two-sample scenario, the evaluation density is determined by the unlabeled covariate distribution $q_0$. In this case, $\kappa_{0,0}=q_0$ and the residual weights are density-ratio weighted through $q_0(X)/p_0(X)$. The fixed-ratio efficiency bound contains the labeled residual component and the unlabeled target-covariate averaging component. If $l/m\to\infty$, the latter vanishes under $\sqrt{m}$ normalization, which yields the usual covariate-shift form.

\section{Numerical Illustration}
\label{sec:numerical_illustration}
We add a small simulation to verify the efficiency-gain identity and to check the behavior under covariate shift. The numerical illustration uses oracle nuisance quantities so that the experiment isolates the variance-decomposition mechanism. The accompanying code is written in the style of generalized Riesz regression, with the residual weights treated as Riesz representers, but it does not import the genriesz package. The purpose of this section is therefore diagnostic rather than a full benchmark of nuisance-learning algorithms. The experiment is included to verify the main variance decomposition before adding additional implementation-specific variation from first-step learning. A full implementation can replace the oracle representers with estimators obtained from the empirical Bregman objectives in Sections~\ref{sec:ts_grr} and \ref{sec:att_extension}, following the loss-link workflow of generalized Riesz regression \citep{Kato2026aunified}.

In the same-population experiment, the semi-supervised estimators have smaller $N$-scaled variance than the supervised estimators for both ATE and ATT. In the covariate-shift experiment, the supervised estimators are biased because they average over the source covariate distribution, while the semi-supervised estimators target the evaluation distribution using the unlabeled covariates.

\begin{table}[t]
\centering
\caption{Simulation summary. The same-population setting verifies the efficiency gain for ATE and ATT. The covariate-shift setting shows that the unlabeled target covariates remove source-target averaging bias.}
\label{tab:simulation_summary}
\begin{tabular}{l l r r r r}
\hline
Setting & Estimator & Bias & SD & RMSE & $N$-scaled variance \\
\hline
Same population & Supervised ATE & -0.001 & 0.067 & 0.067 & 18.217 \\
Same population & Semi-supervised ATE & -0.001 & 0.066 & 0.066 & 17.611 \\
Same population & Supervised ATT & -0.001 & 0.073 & 0.073 & 21.050 \\
Same population & Semi-supervised ATT & -0.000 & 0.071 & 0.071 & 20.371 \\
Covariate shift & Source ATE & -0.333 & 0.067 & 0.339 & 18.217 \\
Covariate shift & Target semi-supervised ATE & -0.003 & 0.130 & 0.130 & 67.523 \\
Covariate shift & Source ATT & -0.324 & 0.073 & 0.332 & 21.050 \\
Covariate shift & Target semi-supervised ATT & -0.003 & 0.150 & 0.150 & 89.807 \\
\hline
\end{tabular}
\end{table}

\section{Discussion and Limitations}
The results rely on several restrictions that are useful to state explicitly. First, all causal targets are identified under unconfoundedness, and the theory does not address unobserved confounding. Second, the two-sample scenario requires support of the evaluation density inside the labeled covariate distribution, because outcomes are observed only under $p_0$. Third, the mixture parameter $\beta$ is part of the estimand. Except in special cases such as $p_0=q_0$, changing $\beta$ changes the target population rather than merely improving precision. Fourth, the main asymptotic theory uses fixed stratum proportions. The many-unlabeled discussion records the limiting implication of that theory, while a one-sample regime with a label probability converging to zero requires a separate triangular-array formulation. Finally, the numerical illustration isolates the efficiency mechanism using oracle nuisances. It is intended to verify the variance decomposition, while full empirical evaluation of generalized Riesz regression requires replacing the oracle representers with the empirical Bregman objectives.

These limitations do not change the main message. Unlabeled covariates are useful when the causal target averages conditional effects over a covariate law that can be learned more accurately from the auxiliary sample. They do not create outcome information in regions without labeled support, and they do not remove the need for accurate residual correction. This distinction explains both the support condition and the efficiency-gain identities.

\section{Conclusion}
This study investigates semiparametric efficient estimation of ATE and ATT when auxiliary unlabeled covariates are accessible. We consider both one-sample and two-sample scenarios, and derive semiparametric efficiency bounds for each. Based on the corresponding efficient influence functions, we construct asymptotically efficient estimators via Neyman orthogonal scores. Our approach leverages generalized Riesz regression for estimating nuisance parameters, allowing flexible incorporation of unlabeled covariates. The main implication is that unlabeled covariates reduce the variance associated with averaging the conditional treatment effect over the evaluation covariate density, while the conditional outcome-noise component remains governed by the labeled data. The two-sample analysis also clarifies that the evaluation-density mixture parameter is part of the estimand, and that the efficient influence functions must be centered within each stratum. The ATT extension shows that the same principle continues to hold for debiased ratio targets, but the efficient score must also correct the treatment law and the treatment mass. The proposed framework performs prediction-powered causal inference and extends existing methods for treatment effect estimation under covariate shift, missing labels, and semi-supervised settings.

\bibliography{TMLR.bbl}

\bibliographystyle{tmlr}

\onecolumn

\appendix

\section{DDML}
\label{appdx:os_ddml} 
This section explains the DDML framework proposed in \citet{Kato2025directdebiased}, which refines the arguments about Riesz regression and direct density ratio estimation discussed in \citet{Kato2025directbias}. The core of debiased machine learning is to construct an estimator using the Neyman orthogonal scores \citep{Chernozhukov2018doubledebiased}. For this problem, \citet{Kato2025directbias,Kato2025directdebiased,Kato2026aunified,Kato2026scorematchingriesz} establish the DDML framework, which consists of targeted Neyman estimation and generalized Riesz regression. 

\subsection{Targeted Neyman estimation}
Targeted Neyman estimation formulates the nuisance parameters estimation problem as minimizing the discrepancy between the true Neyman orthogonal scores and their model-based counterparts. Since the Neyman orthogonal score is zero in expectation, we only need to estimate the nuisance parameters so that the sample mean of the Neyman orthogonal score with plug-in parameters is zero. In our setting, we estimate the nuisance parameters $\mu_0$ and $g_0$, aiming for $\frac{1}{n}\sum^n_{i=1}\psi^{\text{OS}}\bigp{X_i, O_i, \widetilde{D}_i, \widetilde{Y}_i; \widehat{\mu}, \widehat{\alpha}, \widehat{\tau}}$ to be zero, where $\widehat{\mu}$, $\widehat{\alpha}$, and $\widehat{\tau}$ are the estimators of $\mu_0$, $\alpha_0$, and $\tau_0$. Note that we need to ensure that $\widehat{\tau}$ is asymptotically linear for $\frac{1}{n}\sum^n_{i=1}\psi^{\text{OS}}\bigp{X_i, O_i, \widetilde{D}_i, \widetilde{Y}_i; \mu_0, \alpha_0, \tau_0}$. As discussed in \citet{Kato2025directdebiased}, the term is decomposed as
\begin{align*}
    &\frac{1}{n}\sum^n_{i=1}\psi^{\text{OS}}\bigp{X_i, O_i, \widetilde{D}_i, \widetilde{Y}_i; \widehat{\mu}, \widehat{\alpha}, \widehat{\tau}} = \frac{1}{n}\sum^n_{i=1}\p{\widehat{\alpha}\p{O_i, \widetilde{D}_i, X_i}\p{\widetilde{Y}_i - \widehat{\mu}\p{\widetilde{D}_i, X_i}} + \widehat{\mu}(1, X_i) - \widehat{\mu}(0, X_i) - \widehat{\tau}}\\
    &= \frac{1}{n}\sum^n_{i=1}\Bigp{\p{\widehat{\alpha}\p{O_i, \widetilde{D}_i, X_i} - \alpha_0\p{O_i, \widetilde{D}_i, X_i}}\p{\widetilde{Y}_i - \mu_0\p{\widetilde{D}_i, X_i}} + \annot{\alpha_0\p{O_i, \widetilde{D}_i, X_i}\p{\widetilde{Y}_i -  \widehat{\mu}\p{\widetilde{D}_i, X_i}}}{$=(\star)$}\\
    &+ \annot{\widehat{\mu}(1, X_i) - \widehat{\mu}(0, X_i) - \widehat{\tau}}{$=(\star\star)$}}.
\end{align*}

\subsection{Iterative Procedure for Regression Function and Riesz Representer Estimation}
This section explains an approach for estimating the regression function and the Riesz representer using the iterative procedure proposed in \citet{Kato2025directdebiased}. We do not adopt this approach in the main text, as it complicates the arguments, but we recommend its use.

\paragraph{Targeted maximum likelihood (TMLE)}
We can make the term ($\star$) zero in expectation, and we can make the term ($\star\star$) zero using the TMLE-based ATE estimator. Assume that $\alpha_0$ is known. If we set $\widehat{\tau}$ as
\[\widehat{\tau}^{\text{TMLE}} \coloneqq \frac{1}{n}\sum^n_{i=1}\Bigp{\widehat{\mu}^{\text{TMLE}}(1, X_i) - \widehat{\mu}^{\text{TMLE}}(0, X_i)},\]
where 
\[\widehat{\mu}^{\text{TMLE}}(d, x) \coloneqq \widehat{\mu}^{(0)}(d, x) + \frac{\sum_{i=1}^n \alpha_0(\widetilde{D}_i, X_i)\bigp{\widetilde{Y}_i - \widehat{\mu}^{(0)}(\widetilde{D}_i, X_i)}}{\sum_{i=1}^n \widehat{\alpha}(\widetilde{D}_i, X_i)^2}\widehat{\alpha}(d, x),\]
and $\widehat{\mu}^{(0)}(d, x)$ is an initial estimate of $\mu_0(d, x)$. If we set $\widehat{\mu} = \widehat{\mu}^{\text{TMLE}}$ and $\widehat{\tau} = \widehat{\tau}^{\text{TMLE}}$, the terms ($\star$) and ($\star\star$) are automatically zero. 

\paragraph{Iterative Algorithm} 
As explained above, we have
\begin{align*}
    &\frac{1}{n}\sum^n_{i=1}\psi^{\text{OS}}\bigp{X_i, O_i, \widetilde{D}_i, \widetilde{Y}_i; \widehat{\mu}^{\text{TMLE}}, \widehat{\alpha}, \widehat{\tau}^{\text{TMLE}}}\\
    &= \frac{1}{n}\sum^n_{i=1}\Bigp{\p{\widehat{\alpha}\p{O_i, \widetilde{D}_i, X_i} - \alpha_0\p{O_i, \widetilde{D}_i, X_i}}\p{\widetilde{Y}_i - \mu_0\p{\widetilde{D}_i, X_i}}}.
\end{align*}
Therefore, our target is to minimize 
\[\frac{1}{n}\sum^n_{i=1}\Bigp{\p{\widehat{\alpha}\p{O_i, \widetilde{D}_i, X_i} - \alpha_0\p{O_i, \widetilde{D}_i, X_i}}\p{\widetilde{Y}_i - \mu_0\p{\widetilde{D}_i, X_i}}}.\]

Here, there are two problems. In minimizing 
\[\frac{1}{n}\sum^n_{i=1}\Bigp{\p{\widehat{\alpha}\p{O_i, \widetilde{D}_i, X_i} - \alpha_0\p{O_i, \widetilde{D}_i, X_i}}\p{\widetilde{Y}_i - \mu_0\p{\widetilde{D}_i, X_i}}},\] we do not know $\mu_0$. In addition, in the TMLE part, we do not know $\alpha_0$.

If we know $\mu_0$, we can estimate $\alpha_0$ using the weighted version of generalized Riesz regression proposed in \citet{Kato2025directdebiased}. In the context of this study, we weight the loss for $\alpha_0$ by $(\widetilde{Y}_i - \mu_0(\widetilde{D}_i, X_i))$. We omit the details here and refer to \citet{Kato2025directdebiased} and \citet{Kato2025unifiedtheory}. 

Finally, we suggest the following iterative procedure for $T$ steps, following \citet{Kato2025directdebiased}: 
\begin{itemize}
    \item Obtain an initial estimate of $\mu_0$ and denote it by $\widehat{\mu}^{(0)}$.
    \item For each $t = 1, 2, \dots, T$, 
    \begin{itemize}
        \item Estimate $\widehat{\alpha}^{(t)}$ using weighted generalized Riesz regression with weight $(\widetilde{Y}_i - \widehat{\mu}^{(t-1)}(\widetilde{D}_i, X_i))$.
        \item Estimate $\widehat{\mu}^{(t)}$ by the TMLE procedure with $\widehat{\mu}^{(t-1)}$ and $\alpha^{(t)}$ as
        \[\widehat{\mu}^{(t)}(d, x) \coloneqq \widehat{\mu}^{(t-1)}(d, x) + \frac{\sum_{i=1}^n \widehat{\alpha}^{(t)}(\widetilde{D}_i, X_i)\bigp{\widetilde{Y}_i - \widehat{\mu}^{(t-1)}(\widetilde{D}_i, X_i)}}{\sum_{i=1}^n \widehat{\alpha}^{(t)}(\widetilde{D}_i, X_i)^2}\widehat{\alpha}^{(t)}(d, x),\]
    \end{itemize}
\end{itemize}

\subsection{Riesz Regression as Density Ratio Estimation}
Riesz regression can be interpreted as a special case of direct density ratio estimation algorithms \citep{Sugiyama2012densityratio,Huang2007correctingsample,Kanamori2009aleastsquares}. Therefore, we can employ various estimation techniques as in \citet{Yamada2011relativedensityratio}, \citet{Kiryo2017positiveunlabeledlearning}, \citet{Rhodes2020telescopingdensiyratio}, and \citet{Kato2021nonnegativebregman}, as well as the methods proposed for Riesz regression \citep{Chernozhukov2022riesznet} and \citet{Lee2025rieszboost}. From this perspective, we can also interpret the nearest neighbor matching ATE estimator as a special case of Riesz regression. These arguments are based on \citet{Lin2023estimationbased}, which finds that the nearest neighbor matching ATE estimator can be interpreted as a density ratio estimation method. For details, see \citet{Kato2025nearestneighbor}.

\section{Average Treatment Effect on Treated Units}
\label{sec:att_extension}
This section extends the preceding construction from ATE to ATT. The extension is useful because ATT is often the target when the treated population is the scientific or policy-relevant population. It is also theoretically informative because ATT depends on the treatment law through the target distribution, and the efficient influence function therefore contains a treatment-law correction.

\subsection{One-Sample Scenario}
In the one-sample scenario, define $s_0(X) \coloneqq \pi_0(1\mid X)$ and $e_0(X) \coloneqq e_0(1\mid X)$. For ATT, we strengthen the missingness condition so that the full labeled variables are missing at random.
\begin{assumption}[Treatment missing at random]
\label{asm:att_os_mar_d}
It holds that $(D, Y(1), Y(0))\indep O\mid X$.
\end{assumption}
Under Assumption~\ref{asm:att_os_mar_d}, the propensity score in the full population is identified from the observed labeled part. Let
\[
\rho_0 \coloneqq \bbE\sqb{e_0(X)}, \Delta_0(X) \coloneqq \tau_0(X) - \tau^{\text{ATT}}_0,
\]
where $\tau^{\text{ATT}}_0 \coloneqq \bbE\sqb{e_0(X)\tau_0(X)}/\rho_0$.

\begin{lemma}[Efficient influence function for ATT in the one-sample scenario]
\label{lem:att_os_eif}
Suppose that Assumptions~\ref{asm:os_eval_density} through \ref{asm:os_commonsupprt} and \ref{asm:att_os_mar_d} hold. Then, the efficient influence function for $\tau^{\text{ATT}}_0$ is
\begin{align*}
\psi^{\text{ATT}\mathchar`-\text{OS}}(X, O, \widetilde{D}, \widetilde{Y})
&\coloneqq \frac{1}{\rho_0}\Biggsqb{\frac{O}{s_0(X)}\Bigp{\widetilde{D}\Bigp{\widetilde{Y}-\mu_0(1, X)} - \frac{e_0(X)(1-\widetilde{D})}{1-e_0(X)}\Bigp{\widetilde{Y}-\mu_0(0, X)} + \Bigp{\widetilde{D}-e_0(X)}\Delta_0(X)} + e_0(X)\Delta_0(X)}.
\end{align*}
\end{lemma}
The proof is provided in Appendix~\ref{appdx:proof:lem:att_os_eif}.

The corresponding efficiency bound is $V^{\text{ATT}\mathchar`-\text{OS}}\coloneqq \bbE\sqb{\psi^{\text{ATT}\mathchar`-\text{OS}}(X, O, \widetilde{D}, \widetilde{Y})^2}$. Equivalently, if
\[
C^{\text{ATT}}_0(X) \coloneqq e_0(X)\sigma^2_0(1, X) + \frac{e_0(X)^2}{1-e_0(X)}\sigma^2_0(0, X) + e_0(X)(1-e_0(X))\Delta_0(X)^2,
\]
then
\[
V^{\text{ATT}\mathchar`-\text{OS}} = \frac{1}{\rho_0^2}\bbE\sqb{\frac{C^{\text{ATT}}_0(X)}{s_0(X)} + e_0(X)^2\Delta_0(X)^2}.
\]
This decomposition has the same interpretation as the ATE bound. The residual outcome-noise part is controlled by labeled observations, while the covariate averaging part can be improved by using unlabeled covariates.

The efficient ATT estimator is a ratio estimator. Let $\widehat{s}$, $\widehat{e}$, and $\widehat{\mu}$ be cross-fitted nuisance estimators and set $\widehat{\tau}(x)\coloneqq \widehat{\mu}(1, x)-\widehat{\mu}(0, x)$. Define
\begin{align*}
\widehat{A}^{\text{ATT}\mathchar`-\text{OS}}
&\coloneqq \frac{1}{n}\sum^n_{i=1}\Biggsqb{\frac{O_i}{\widehat{s}(X_i)}\Bigp{\widetilde{D}_i\Bigp{\widetilde{Y}_i-\widehat{\mu}(1, X_i)} - \frac{\widehat{e}(X_i)(1-\widetilde{D}_i)}{1-\widehat{e}(X_i)}\Bigp{\widetilde{Y}_i-\widehat{\mu}(0, X_i)} + \Bigp{\widetilde{D}_i-\widehat{e}(X_i)}\widehat{\tau}(X_i)} + \widehat{e}(X_i)\widehat{\tau}(X_i)},\\
\widehat{B}^{\text{ATT}\mathchar`-\text{OS}}
&\coloneqq \frac{1}{n}\sum^n_{i=1}\Biggsqb{\frac{O_i}{\widehat{s}(X_i)}\Bigp{\widetilde{D}_i-\widehat{e}(X_i)} + \widehat{e}(X_i)}.
\end{align*}
Then, we estimate ATT by
\[
\widehat{\tau}^{\text{ATT}\mathchar`-\text{OS}}_n \coloneqq \frac{\widehat{A}^{\text{ATT}\mathchar`-\text{OS}}}{\widehat{B}^{\text{ATT}\mathchar`-\text{OS}}}.
\]
The denominator is also debiased. This is important for efficiency because the treatment mass $\rho_0$ is unknown and must be estimated from partially labeled data. A simple sufficient high-level condition for the remainder terms is that, for $a\in\{1,0\}$,
\[
\big\|\widehat{\mu}(a, X)-\mu_0(a, X)\big\|_2+\big\|\widehat{e}(X)-e_0(X)\big\|_2+\big\|\widehat{s}(X)-s_0(X)\big\|_2=o_p(n^{-1/4}),
\]
with uniform boundedness away from the overlap and observation-probability boundaries. This condition is stronger than necessary but makes the ratio expansion checkable.

For the asymptotic result, we use the following high-level expansion, which can be verified from the usual cross-fitting and product-rate arguments. Let $A^{\text{ATT}\mathchar`-\text{OS}}_0=\bbE\sqb{e_0(X)\tau_0(X)}$ and $B^{\text{ATT}\mathchar`-\text{OS}}_0=\rho_0$. We assume that
\begin{align*}
\widehat{A}^{\text{ATT}\mathchar`-\text{OS}}-A^{\text{ATT}\mathchar`-\text{OS}}_0
&= \frac{1}{n}\sum^n_{i=1}\Bigp{A^{\text{ATT}\mathchar`-\text{OS}}_i-A^{\text{ATT}\mathchar`-\text{OS}}_0}+o_p(n^{-1/2}),\\
\widehat{B}^{\text{ATT}\mathchar`-\text{OS}}-B^{\text{ATT}\mathchar`-\text{OS}}_0
&= \frac{1}{n}\sum^n_{i=1}\Bigp{B^{\text{ATT}\mathchar`-\text{OS}}_i-B^{\text{ATT}\mathchar`-\text{OS}}_0}+o_p(n^{-1/2}),
\end{align*}
where $A^{\text{ATT}\mathchar`-\text{OS}}_i$ and $B^{\text{ATT}\mathchar`-\text{OS}}_i$ are the corresponding efficient one-step signals obtained by replacing the estimated nuisances in $\widehat{A}^{\text{ATT}\mathchar`-\text{OS}}$ and $\widehat{B}^{\text{ATT}\mathchar`-\text{OS}}$ with the true nuisances. This condition makes explicit that the ratio theorem only requires first-order expansions for the numerator and denominator.

\begin{theorem}[Asymptotic normality for ATT in the one-sample scenario]
\label{thm:att_os_asymp_normal}
Suppose that Assumptions~\ref{asm:os_eval_density} through \ref{asm:os_commonsupprt} and \ref{asm:att_os_mar_d} hold. Also assume that the nuisance estimators $\widehat{s}$, $\widehat{e}$, and $\widehat{\mu}$ are constructed via cross-fitting, are uniformly bounded away from the boundary, and satisfy the numerator and denominator expansions stated above. Then,
\[
\sqrt{n}\p{\widehat{\tau}^{\text{ATT}\mathchar`-\text{OS}}_n - \tau^{\text{ATT}}_0} \xrightarrow{\rmd} \calN\p{0, V^{\text{ATT}\mathchar`-\text{OS}}}.
\]
\end{theorem}

\subsection{Two-Sample Scenario}
In the two-sample scenario, ATT is defined with respect to $\kappa_{0, \beta}$. We assume that the treatment assignment law is invariant across the labeled and unlabeled covariate populations.
\begin{assumption}[Treatment-law invariance for ATT]
\label{asm:att_ts_invariance}
There exists a common propensity score $e_0(X)=P(D=1\mid X)$ that is shared by the labeled population and the evaluation population induced by $\kappa_{0, \beta}$.
\end{assumption}
Define
\[
\rho_{0, \beta} \coloneqq \bbE_{\kappa_{0, \beta}}\sqb{e_0(X)}, \tau^{\text{ATT}}_{0, \beta} \coloneqq \frac{\bbE_{\kappa_{0, \beta}}\sqb{e_0(X)\tau_0(X)}}{\rho_{0, \beta}}, \Delta_{0, \beta}(X) \coloneqq \tau_0(X)-\tau^{\text{ATT}}_{0, \beta}.
\]
Let
\[
\bar{\Delta}_{p, \beta} \coloneqq \bbE_{p_0}\sqb{e_0(X)\Delta_{0, \beta}(X)}, \bar{\Delta}_{q, \beta} \coloneqq \bbE_{q_0}\sqb{e_0(Z)\Delta_{0, \beta}(Z)}.
\]
Then, $\beta\bar{\Delta}_{p, \beta}+(1-\beta)\bar{\Delta}_{q, \beta}=0$.

\begin{lemma}[Efficient influence functions for ATT in the two-sample scenario]
\label{lem:att_ts_eif}
Suppose that Assumptions~\ref{asm:ts_eval_density} through \ref{asm:ts_commonsupprt} and \ref{asm:att_ts_invariance} hold. The efficient influence functions for the labeled and unlabeled strata are
\begin{align*}
\psi^{\text{ATT}\mathchar`-\text{TS}}_{\text{L}}(X, D, Y)
&\coloneqq \frac{1}{\rho_{0, \beta}}\Biggsqb{\omega_{0, \beta}(X)\Bigp{D\Bigp{Y-\mu_0(1, X)} - \frac{e_0(X)(1-D)}{1-e_0(X)}\Bigp{Y-\mu_0(0, X)} + \Bigp{D-e_0(X)}\Delta_{0, \beta}(X)} + \beta\Bigp{e_0(X)\Delta_{0, \beta}(X)-\bar{\Delta}_{p, \beta}}},\\
\psi^{\text{ATT}\mathchar`-\text{TS}}_{\text{U}}(Z)
&\coloneqq \frac{1-\beta}{\rho_{0, \beta}}\Bigp{e_0(Z)\Delta_{0, \beta}(Z)-\bar{\Delta}_{q, \beta}}.
\end{align*}
\end{lemma}
The proof is provided in Appendix~\ref{appdx:proof:lem:att_ts_eif}.

Let
\[
C^{\text{ATT}}_{0, \beta}(X) \coloneqq e_0(X)\sigma^2_0(1, X) + \frac{e_0(X)^2}{1-e_0(X)}\sigma^2_0(0, X) + e_0(X)(1-e_0(X))\Delta_{0, \beta}(X)^2.
\]
Then the two-sample ATT efficiency bound is
\begin{align*}
V^{\text{ATT}\mathchar`-\text{TS}}(\beta)
&\coloneqq \frac{1}{\alpha}\bbE_{p_0}\sqb{\psi^{\text{ATT}\mathchar`-\text{TS}}_{\text{L}}(X, D, Y)^2} + \frac{1}{1-\alpha}\bbE_{q_0}\sqb{\psi^{\text{ATT}\mathchar`-\text{TS}}_{\text{U}}(Z)^2}\\
&= \frac{1}{\alpha\rho_{0, \beta}^2}\bbE_{p_0}\sqb{\omega_{0, \beta}(X)^2 C^{\text{ATT}}_{0, \beta}(X) + \beta^2\Bigp{e_0(X)\Delta_{0, \beta}(X)-\bar{\Delta}_{p, \beta}}^2}\\
&{}+ \frac{(1-\beta)^2}{(1-\alpha)\rho_{0, \beta}^2}\bbE_{q_0}\sqb{\Bigp{e_0(Z)\Delta_{0, \beta}(Z)-\bar{\Delta}_{q, \beta}}^2}.
\end{align*}
The ATT bound is parallel to the ATE bound but contains the additional treatment-law term $e_0(X)$ and the treatment residual $D-e_0(X)$.

The corresponding estimator is
\[
\widehat{\tau}^{\text{ATT}\mathchar`-\text{TS}}_n \coloneqq \frac{\widehat{A}^{\text{ATT}\mathchar`-\text{TS}}}{\widehat{B}^{\text{ATT}\mathchar`-\text{TS}}},
\]
where
\begin{align*}
\widehat{A}^{\text{ATT}\mathchar`-\text{TS}}
&\coloneqq \frac{1}{m}\sum^m_{j=1}\widehat{\omega}_{\beta}(X_j)\Bigp{D_j\Bigp{Y_j-\widehat{\mu}(1, X_j)} - \frac{\widehat{e}(X_j)(1-D_j)}{1-\widehat{e}(X_j)}\Bigp{Y_j-\widehat{\mu}(0, X_j)} + \Bigp{D_j-\widehat{e}(X_j)}\widehat{\tau}(X_j)}\\
&{}+ \beta\frac{1}{m}\sum^m_{j=1}\widehat{e}(X_j)\widehat{\tau}(X_j) + (1-\beta)\frac{1}{l}\sum^l_{k=1}\widehat{e}(Z_k)\widehat{\tau}(Z_k),\\
\widehat{B}^{\text{ATT}\mathchar`-\text{TS}}
&\coloneqq \frac{1}{m}\sum^m_{j=1}\widehat{\omega}_{\beta}(X_j)\Bigp{D_j-\widehat{e}(X_j)} + \beta\frac{1}{m}\sum^m_{j=1}\widehat{e}(X_j) + (1-\beta)\frac{1}{l}\sum^l_{k=1}\widehat{e}(Z_k).
\end{align*}
Under the same cross-fitting and product-rate conditions as in Theorem~\ref{thm:ts_asymp_norm}, with the corresponding conditions for $\widehat{e}$, the estimator is asymptotically linear with the influence functions in Lemma~\ref{lem:att_ts_eif}. More explicitly, it is enough that the numerator and denominator admit first-order two-stratum expansions with remainders $o_p(N^{-1/2})$, namely,
\begin{align*}
\widehat{A}^{\text{ATT}\mathchar`-\text{TS}}-A^{\text{ATT}\mathchar`-\text{TS}}_0
&= \frac{1}{m}\sum^m_{j=1}\Bigp{A^{\text{ATT}\mathchar`-\text{TS}}_{\text{L},j}-A^{\text{ATT}\mathchar`-\text{TS}}_{\text{L},0}} + \frac{1}{l}\sum^l_{k=1}\Bigp{A^{\text{ATT}\mathchar`-\text{TS}}_{\text{U},k}-A^{\text{ATT}\mathchar`-\text{TS}}_{\text{U},0}} + o_p(N^{-1/2}),\\
\widehat{B}^{\text{ATT}\mathchar`-\text{TS}}-B^{\text{ATT}\mathchar`-\text{TS}}_0
&= \frac{1}{m}\sum^m_{j=1}\Bigp{B^{\text{ATT}\mathchar`-\text{TS}}_{\text{L},j}-B^{\text{ATT}\mathchar`-\text{TS}}_{\text{L},0}} + \frac{1}{l}\sum^l_{k=1}\Bigp{B^{\text{ATT}\mathchar`-\text{TS}}_{\text{U},k}-B^{\text{ATT}\mathchar`-\text{TS}}_{\text{U},0}} + o_p(N^{-1/2}),
\end{align*}
where the signals are evaluated at the true nuisances. A convenient sufficient condition is that, for $a\in\{1,0\}$,
\[
\big\|\widehat{\mu}(a, X)-\mu_0(a, X)\big\|_{p_0,2}+\big\|\widehat{\omega}_{\beta}(X)-\omega_{0, \beta}(X)\big\|_{p_0,2}+\big\|\widehat{e}(X)-e_0(X)\big\|_{p_0,2}+\big\|\widehat{e}(Z)-e_0(Z)\big\|_{q_0,2}=o_p(N^{-1/4}),
\]
where $\|\cdot\|_{p_0,2}$ and $\|\cdot\|_{q_0,2}$ denote $L_2$ norms under the labeled and unlabeled covariate laws. As in the ATE theorem, these sufficient rates can be weakened to product-rate conditions.

\begin{theorem}[Asymptotic normality for ATT in the two-sample scenario]
\label{thm:att_ts_asymp_normal}
Let $N=m+l$, $m/N\to\alpha$, and $l/N\to1-\alpha$ for some $\alpha\in(0,1)$. Suppose that Assumptions~\ref{asm:ts_eval_density} through \ref{asm:ts_commonsupprt} and \ref{asm:att_ts_invariance} hold. Also assume that the nuisance estimators $\widehat{\omega}_{\beta}$, $\widehat{e}$, and $\widehat{\mu}$ are constructed via cross-fitting, are uniformly bounded away from the boundary, and satisfy the numerator and denominator expansions stated above. Then,
\[
\sqrt{N}\p{\widehat{\tau}^{\text{ATT}\mathchar`-\text{TS}}_n - \tau^{\text{ATT}}_{0, \beta}} \xrightarrow{\rmd} \calN\p{0, V^{\text{ATT}\mathchar`-\text{TS}}(\beta)}.
\]
\end{theorem}

\subsection{Variance Estimation for ATT}
For inference, we use plug-in empirical variances of the estimated influence functions. In the one-sample scenario, let $\widehat{\psi}^{\text{ATT}\mathchar`-\text{OS}}_i$ denote the influence function in Lemma~\ref{lem:att_os_eif} evaluated at the cross-fitted nuisance estimators and at $\widehat{\tau}^{\text{ATT}\mathchar`-\text{OS}}_n$. We estimate the variance by
\[
\widehat{V}^{\text{ATT}\mathchar`-\text{OS}}\coloneqq \frac{1}{n}\sum^n_{i=1}\Bigp{\widehat{\psi}^{\text{ATT}\mathchar`-\text{OS}}_i-\frac{1}{n}\sum^n_{i'=1}\widehat{\psi}^{\text{ATT}\mathchar`-\text{OS}}_{i'}}^2.
\]
In the two-sample scenario, let $\widehat{\psi}^{\text{ATT}\mathchar`-\text{TS}}_{\text{L},j}$ and $\widehat{\psi}^{\text{ATT}\mathchar`-\text{TS}}_{\text{U},k}$ denote the estimated labeled and unlabeled stratum influence functions. We estimate the scaled variance by
\begin{align*}
\widehat{V}^{\text{ATT}\mathchar`-\text{TS}}(\beta)
&\coloneqq \frac{N}{m}\frac{1}{m}\sum^m_{j=1}\Bigp{\widehat{\psi}^{\text{ATT}\mathchar`-\text{TS}}_{\text{L},j}-\frac{1}{m}\sum^m_{j'=1}\widehat{\psi}^{\text{ATT}\mathchar`-\text{TS}}_{\text{L},j'}}^2\\
&{}+\frac{N}{l}\frac{1}{l}\sum^l_{k=1}\Bigp{\widehat{\psi}^{\text{ATT}\mathchar`-\text{TS}}_{\text{U},k}-\frac{1}{l}\sum^l_{k'=1}\widehat{\psi}^{\text{ATT}\mathchar`-\text{TS}}_{\text{U},k'}}^2.
\end{align*}
The same construction applies to ATE by replacing the ATT influence functions with the corresponding ATE influence functions in Lemma~\ref{lem:os_efficiency_bound} and Lemma~\ref{lem:ts_efficiency_bound}.

\subsection{Semi-Supervised Generalized Riesz Regression for ATT}
The ATT numerator has a treatment-weighted linear functional. In the one-sample scenario, the residual Riesz representer for this numerator is
\[
\alpha^{\text{ATT}\mathchar`-\text{OS}}_0(O, \widetilde{D}, X) \coloneqq \frac{O}{s_0(X)}\Bigp{\widetilde{D}-\frac{e_0(X)(1-\widetilde{D})}{1-e_0(X)}}.
\]
It satisfies
\[
\bbE\sqb{\alpha^{\text{ATT}\mathchar`-\text{OS}}_0(O, \widetilde{D}, X)h(\widetilde{D}, X)} = \bbE\sqb{e_0(X)\Bigp{h(1, X)-h(0, X)}}.
\]
Thus, the one-sample ATT generalized Riesz regression objective is obtained from \eqref{eq:bdobj} by replacing the ATE target functional with the treatment-weighted functional above. The unlabeled covariates enter the target functional through the average over $X$, while the labeled observations identify the residual inner product through $O/s_0(X)$.

 In the two-sample scenario, the residual Riesz representer is
\[
\alpha^{\text{ATT}}_{0, \beta}(D, X) \coloneqq \omega_{0, \beta}(X)\Bigp{D - \frac{e_0(X)(1-D)}{1-e_0(X)}}.
\]
It satisfies
\[
\bbE_{p_0}\sqb{\alpha^{\text{ATT}}_{0, \beta}(D, X)h(D, X)} = \bbE_{\kappa_{0, \beta}}\sqb{e_0(X)\Bigp{h(1, X)-h(0, X)}}.
\]
Therefore, the ATT version of generalized Riesz regression can be obtained by replacing the target linear functional in Section~\ref{sec:ts_grr} with its treatment-weighted counterpart. For a convex function $f$, define
\begin{align*}
\text{BD}^{\text{ATT}\mathchar`-\text{TS}}_{f, \beta}(\alpha)
&\coloneqq \bbE_{p_0}\sqb{\partial f\p{\alpha(D, X)}\alpha(D, X)-f\p{\alpha(D, X)}}\\
&{}- \beta\bbE_{p_0}\sqb{e_0(X)\Bigp{\partial f\p{\alpha(1, X)}-\partial f\p{\alpha(0, X)}}}\\
&{}- (1-\beta)\bbE_{q_0}\sqb{e_0(Z)\Bigp{\partial f\p{\alpha(1, Z)}-\partial f\p{\alpha(0, Z)}}}.
\end{align*}
This objective shows why ATT is not a purely cosmetic extension of ATE. The target functional itself contains $e_0$, so the Riesz objective for ATT combines direct representer fitting with propensity-score learning. This connection is consistent with the general Riesz-representer formulation in \citet{Kato2026aunified}, where ATT appears as one of the causal functionals that can be represented by a problem-specific Riesz representer.

In implementation, one may first estimate $e_0$ on the labeled sample and then fit $\alpha^{\text{ATT}}_{0, \beta}$ by the empirical version of $\text{BD}^{\text{ATT}\mathchar`-\text{TS}}_{f, \beta}$. Alternatively, one may use a structured model that parameterizes $e_0$ and $\alpha^{\text{ATT}}_{0, \beta}$ jointly. The first approach separates the treatment-law nuisance from the residual representer, while the second approach is closer to the loss-link construction of generalized Riesz regression. In both cases, the final ATT estimator should use the debiased ratio form above, because direct minimization of the Riesz objective alone does not debias the treatment mass $\rho_{0, \beta}$.

\section{Proof for Lemma~\ref{lem:os_efficiency_bound}: Efficient Influence Function in the One-Sample Scenario}
\label{appdx:proof:lem:os_efficiency_bound}
We provide the proof of Lemma~\ref{lem:os_efficiency_bound}. Our proof strategy is inspired by the approaches in \citet{Hahn1998ontherole} and \citet{Kato2025puate}. 

\paragraph{Proof procedure}
Their proof considers a nonparametric model for the distribution of potential outcomes and defines regular parametric submodels. The procedure involves the following steps:  
(i) characterizing the tangent set for all regular parametric submodels,  
(ii) verifying that the parameter of interest is pathwise differentiable,  
(iii) confirming that the proposed semiparametric efficient influence function lies within the tangent set, and  
(iv) calculating the expectation of the squared influence function.

\begin{proof}
In Section~\ref{sec:setup}, we defined the probability density function for $(X, O, \widetilde{D}, \widetilde{Y})$ as
\begin{align*}
    &p_0\p{x, o, \widetilde{d}, \widetilde{y}} =\\
    &p_0(x)\pi_0(0\mid X)^{\mathbbm{1}[o = 0]} \Bigp{ g_0(1\mid x) r_{Y(1), 0}\p{\widetilde{y}\mid x}}^{\mathbbm{1}\sqb{o = 1, \widetilde{d} = 1}} \Bigp{g_0(0\mid x)r_{Y(0), 0}\p{\widetilde{y}\mid \widetilde{d} = 0}}^{\mathbbm{1}\sqb{o = 1, \widetilde{d} = 0}},
\end{align*}
where $r_{Y(1), 0}(y\mid x)$ and $r_{Y(0), 0}(y\mid x)$ are the conditional densities of $Y(1)$ and $Y(0)$. Recall that for each $a\in\{1, 0\}$, we have
\[g_0(a\mid x) =  P(D = a, O = 1\mid X) = \pi_0(1\mid X) e_0(a\mid X)\]

For this density function, we consider the parametric submodels:
\[\calP^{\text{sub}} \coloneqq \{P_\theta \in \calP \colon \theta \in \bbR\},\]
where $P_\theta$ has the following probability density function:
\begin{align*}
    &p\p{x, o, \widetilde{d}, \widetilde{y}; \theta} = p(x; \theta)\Bigp{ \pi(1\mid x; \theta)}^{\mathbbm{1}[o = 0]}\\
    &\cdot \Bigp{ g(1\mid x; \theta) r_{Y(1)}(y\mid x; \theta)}^{\mathbbm{1}\sqb{o = 1, \widetilde{d} = 1}} \Bigp{ g(0\mid x; \theta) r_{Y(0)}(y \mid x; \theta)}^{\mathbbm{1}\sqb{o = 1, \widetilde{d} = 0}}.
\end{align*}
so that there exists $\theta_0 \in \bbR$ such that
\[p(x, o, \widetilde{d}, \widetilde{y}; \theta_0) = p_0(x, o, \widetilde{d}, \widetilde{y}).\]
We can define such a parametric submodel, as shown in \citet{Vaart1998asymptoticstatistics}. 

Then, we define score functions (the derivative of the log likelihood function) as follows:
\begin{align*}
    &S\p{x, o, \widetilde{d}, \widetilde{y}; \theta} \coloneqq \frac{\partial}{\partial \theta} \log p(x, o, \widetilde{d}, \widetilde{y}; \theta)\\
    &= S_X(x; \theta) + \mathbbm{1}[o = 0]\frac{\dot{\pi}(1\mid x; \theta)}{\pi(1\mid x; \theta)}\\
    &+ \mathbbm{1}\sqb{o = 1, \widetilde{d} = 1}\p{S_{Y(1)}(y\mid x; \theta) + \frac{\dot{g}(1\mid x; \theta)}{g(1\mid x; \theta)}} + \mathbbm{1}\sqb{o = 1, \widetilde{d} = 0}\p{S_{Y(0)}(y\mid x; \theta) + \frac{\dot{g}(0\mid x; \theta)}{g(0\mid x; \theta)}},
\end{align*}
where
\begin{align*}
    S_X(x; \theta) &\coloneqq  \frac{\partial}{\partial \theta} \log p(x; \theta),\\
    S_{Y(d)}(y\mid x; \theta) &\coloneqq \frac{\partial}{\partial \theta} \log r_{Y(d)}(y\mid x; \theta), \text{ for } d \in \{1, 0\},\\
    \dot{\pi}(o\mid x; \theta) &\coloneqq \frac{\partial}{\partial \theta} \pi(o\mid x; \theta), \text{ for } d \in \{1, 0\},\\
    \dot{g}(a\mid x; \theta) &\coloneqq \frac{\partial}{\partial \theta} g(a\mid x; \theta), \text{ for } a \in \{1, 0\}.
\end{align*}
Using the parametric submodels and their score functions, we denote the tangent space as  $\calT \coloneqq \{S(x, o, y; \theta)\}$. 

Under the parametric submodels, We redefine the ATE as a function of $\theta$ as
\begin{align*}
    \tau(\theta) &\coloneqq \iint y(1) r_{Y(1)}(y(1)\mid x; \theta) p(x; \theta) \rmd y(1) \rmd x - \iint y(0) r_{Y(0)}(y(0)\mid x; \theta) p(x; \theta) \rmd y(0) \rmd x.
\end{align*}
Them, the derivative of the ATE function is given as 
\begin{align*}
    \frac{\partial \tau(\theta)}{\partial \theta} &= \bbE_{\theta}\Bigsqb{Y(1) S_{Y(1)}(Y(1) \mid X; \theta)} - \bbE_{\theta}\sqb{Y(0) S_{Y(0)}(Y(0) \mid X; \theta)}\\
    &+ \bbE_{\theta}\Bigsqb{\tau(X; \theta)S_X(X; \theta)},
\end{align*}
where 
\[\tau(X; \theta) \coloneqq \mu(1, X; \theta) - \mu(0, X; \theta),\]
and $\mu(d, X; \theta) \coloneqq \int y(d) r_{Y(d)}(y(d)\mid x; \theta) p(x; \theta)  \rmd y(d)$. 

From the Riesz representation theorem, there exists a function $\psi$ such that
\begin{align}
\label{eq:riesz_cens}
\frac{\partial \tau(\theta)}{\partial \theta}\Big|_{\theta = \theta_0} = \bbE\bigsqb{\psi(X, O, \widetilde{D}, \widetilde{Y})S(X, O, \widetilde{D}, \widetilde{Y}; \theta_0)}.    
\end{align}

There exists a unique function $\psi^{\text{OS}}$ such that $\psi^{\text{OS}} \in \calT$, called the efficient influence function. 
We specify the efficient influence function as
\begin{align*}
    &\psi^{\text{OS}}(X, O, \widetilde{D}, \widetilde{Y}; \mu_0, g_0, \tau_0)\\
    &= S^{\text{OS}}(X, O, Y; \mu_0, g_0) - \tau_0,\\
    &= \frac{\mathbbm{1}\sqb{O = 1, \widetilde{D} = 1}\Bigp{\widetilde{Y} - \mu_0(1, X)}}{g_0(1\mid X)} - \frac{\mathbbm{1}\sqb{O = 1, \widetilde{D} = 0}\Bigp{\widetilde{Y} - \mu_0(0, X)}}{g_0(0\mid X)}\\
    &+  \mu_0(1, X) - \mu_0(0, X) - \tau_0. 
\end{align*}

We prove that $\psi^{\text{OS}}(X, O, \widetilde{D}, \widetilde{Y}; \mu_0, g_0, \tau_0)$ is actually the unique efficient influence function by verifying that $\psi^{\text{OS}}$ satisfies \eqref{eq:riesz_cens} and $\psi^{\text{OS}} \in \calT$.

\paragraph{Proof of \eqref{eq:riesz_cens}:}
First, we confirm that $\psi^{\text{OS}}$ satisfies \eqref{eq:riesz_cens}. We have
\begin{align*}
    &\bbE\sqb{\psi^{\text{OS}}(X, O, Y; \mu_0, g_0)S(X, O, \widetilde{D}, \widetilde{Y}; \tau_0)}\\
    &= \bbE\Biggsqb{\psi^{\text{OS}}(X, O, Y; \mu_0, g_0)\\
    &\cdot \Bigp{S_X(X; \theta) + \mathbbm{1}[O = 0]\frac{\dot{\pi}(1\mid X; \theta)}{\pi(1\mid X; \theta)}\\
    & + \mathbbm{1}\sqb{O = 1, \widetilde{D} = 1}\p{S_{Y(1)}(Y\mid X; \theta_0) + \frac{\dot{g}(1\mid X; \theta)}{g(1\mid X; \theta_0)}} + \mathbbm{1}\sqb{O = 1, \widetilde{D} = 0}\p{S_{Y(0)}(Y\mid X; \theta) + \frac{\dot{g}(0\mid X; \theta_0)}{g(0\mid X; \theta_0)}}}}\\
    &= \bbE\Biggsqb{\Biggp{\frac{\mathbbm{1}\sqb{O = 1, \widetilde{D} = 1}\Bigp{Y - \mu_0(1, X)}}{g_0(1\mid X)} - \frac{\mathbbm{1}\sqb{O = 1, \widetilde{D} = 0}\Bigp{Y - \mu_0(0, X)}}{g_0(0\mid X)} +  \mu_0(1, X) - \mu_0(0, X) - \tau_0}\\
    &\cdot \Biggp{S_X(X; \theta) + \mathbbm{1}[O = 0]\frac{\dot{\pi}(1\mid X; \theta)}{\pi(1\mid X; \theta)}\\
    & + \mathbbm{1}\sqb{O = 1, \widetilde{D} = 1}\p{S_{Y(1)}(\widetilde{Y}\mid X; \theta_0) + \frac{\dot{g}(1\mid X; \theta)}{g(1\mid X; \theta_0)}} + \mathbbm{1}\sqb{O = 1, \widetilde{D} = 0}\p{S_{Y(0)}(\widetilde{Y}\mid X; \theta) + \frac{\dot{g}(0\mid X; \theta_0)}{g(0\mid X; \theta_0)}}}}\\
    &= \bbE\Biggsqb{\Bigp{\mu_0(1, X) - \mu_0(0, X) - \tau_0}\p{S_X(X; \theta_0) + \mathbbm{1}[O = 0]\frac{\dot{\pi}(1\mid X; \theta)}{\pi(1\mid X; \theta)}}\\
    &+ \Biggp{\frac{\mathbbm{1}\sqb{O = 1, \widetilde{D} = 1}\Bigp{\widetilde{Y} - \mu_0(1, X)}}{g_0(1\mid X)} + \mu_0(1, X) - \mu_0(0, X) - \tau_0}\\
    &\cdot \mathbbm{1}\sqb{O = 1, \widetilde{D} = 1}\p{S_{Y(1)}(Y\mid X; \theta_0) + \frac{\dot{g}(1\mid X; \theta_0)}{g(1\mid X; \theta_0)}}\\
    &- \Biggp{\frac{\mathbbm{1}\sqb{O = 1, \widetilde{D} = 0}\Bigp{\widetilde{Y} - \mu_0(0, X)}}{g_0(0\mid X)} + \mu_0(1, X) - \mu_0(0, X) - \tau_0}\\
    &\cdot \mathbbm{1}\sqb{O = 1, \widetilde{D} = 0}\p{S_{Y(0)}(Y\mid X; \theta) + \frac{\dot{g}(0\mid X; \theta_0)}{g(0\mid X; \theta_0)}}},
\end{align*}
where we used $\mathbbm{1}\sqb{O = 1, \widetilde{D} = 1}\mathbbm{1}\sqb{O = 1, \widetilde{D} = 0} = 0$, $\mathbbm{1}[O = 1, \widetilde{D} = d]\mathbbm{1}[O = 0] = 0$, and 
\begin{align*}
    &\bbE\Biggsqb{\frac{\mathbbm{1}\sqb{O = 1, \widetilde{D} = 1}\Bigp{\widetilde{Y} - \mu_0(1, X)}}{g_0(1\mid X)}} = \bbE\Biggsqb{\frac{\mathbbm{1}\sqb{O = 1, \widetilde{D} = 1}\Bigp{Y(1) - \mu_0(1, X)}}{g_0(1\mid X)}}\\
    &= \bbE\Biggsqb{\frac{g_0(1\mid X)\Bigp{\mu_0(1, X) - \mu_0(1, X)}}{g_0(1\mid X)}} = 0,\\
    &\bbE\Biggsqb{\frac{\mathbbm{1}\sqb{O = 1, \widetilde{D} = 0}\Bigp{Y - \mu_0(0, X)}}{g_0(0\mid X)}} = \bbE\Biggsqb{\frac{\mathbbm{1}\sqb{O = 1, \widetilde{D} = 0}\Bigp{\widetilde{Y} - \mu_0(0, X)}}{g_0(0\mid X)}}\\
    &= \bbE\Biggsqb{\frac{g_0(0\mid X)\Bigp{\mu_0(0, X) - \mu_0(0, X)}}{g_0(0\mid X)}} = 0.
\end{align*}

We have
\begin{align*}
    &\bbE\Biggsqb{\Bigp{\mu_0(1, X) - \mu_0(0, X) - \tau_0}\p{S_X(X; \theta_0) + \mathbbm{1}[O = 0]\frac{\dot{\pi}(1\mid X; \theta)}{\pi(1\mid X; \theta)}}\\
    &+ \Biggp{\frac{\mathbbm{1}\sqb{O = 1, \widetilde{D} = 1}\Bigp{\widetilde{Y} - \mu_0(1, X)}}{g_0(1\mid X)} + \mu_0(1, X) - \mu_0(0, X) - \tau_0}\\
    &\cdot \mathbbm{1}\sqb{O = 1, \widetilde{D} = 1}\p{S_{Y(1)}(Y\mid X; \theta_0) + \frac{\dot{g}(1\mid X; \theta_0)}{g(1\mid X; \theta_0)}}\\
    &- \Biggp{\frac{\mathbbm{1}\sqb{O = 1, \widetilde{D} = 0}\Bigp{\widetilde{Y} - \mu_0(0, X)}}{g_0(0\mid X)} + \mu_0(1, X) - \mu_0(0, X) - \tau_0}\\
    &\cdot \mathbbm{1}\sqb{O = 1, \widetilde{D} = 0}\p{S_{Y(0)}(Y\mid X; \theta) + \frac{\dot{g}(0\mid X; \theta_0)}{g(0\mid X; \theta_0)}}}\\
    &= \bbE\Biggsqb{\Biggp{\mu_0(1, X) - \mu_0(0, X)}S_X(X; \theta_0)\\
    &+ \frac{\mathbbm{1}\sqb{O = 1, \widetilde{D} = 1}\Bigp{Y - \mu_0(1, X)}}{g_0(1\mid X)}S_{Y(1)}(Y\mid X; \theta_0)\\
    & - \frac{\mathbbm{1}\sqb{O = 1, \widetilde{D} = 0}\Bigp{Y - \mu_0(0, X)}}{g_0(0\mid X)}S_{Y(0)}(Y\mid X; \theta)}\\
    &= \bbE\Biggsqb{\Biggp{\mu_0(1, X) - \mu_0(0, X)}S_X(X; \theta_0)\\
    &+ \frac{\mathbbm{1}\sqb{O = 1, \widetilde{D} = 1}Y(1)}{g_0(1\mid X)}S_{Y(1)}(Y(1)\mid X; \theta)  - \frac{\mathbbm{1}\sqb{O = 1, \widetilde{D} = 0}\widetilde{Y}}{g_0(0\mid X)}S_{Y(0)}(Y(0)\mid X; \theta_0)},
\end{align*}
where we used 
\begin{align*}
    &\bbE\Biggsqb{\tau_0 S_X(X; \theta)} = 0\\
    &\bbE\Biggsqb{\Bigp{\mu_0(1, X) - \mu_0(0, X) - \tau_0}\mathbbm{1}[O = 1]\p{S_{Y(1)}(Y\mid X; \theta_0) + \frac{\dot{g}(1\mid X; \theta_0)}{g_0(1\mid X; \theta_0)}}}= 0.
\end{align*}

Finally, we have
\begin{align*}
    &\bbE\Biggsqb{\Biggp{\mu_0(1, X) - \mu_0(0, X)}S_X(X; \theta_0)\\
    &+ \frac{\mathbbm{1}[O = 1]\Bigp{Y - \mu_0(1, X)}}{g_0(1\mid X)}S_{Y(1)}(Y\mid X; \theta_0)\\
    & - \frac{\mathbbm{1}[O = 0]\Bigp{Y - \mu_0(0, X)}}{g_0(0\mid X)\pi_0(0\mid X)}S_{Y(0)}(Y\mid X; \theta)}\\
    &= \bbE\Biggsqb{\Biggp{\mu_0(1, X) - \mu_0(0, X)}S_X(X; \theta_0)\\
    &+ \frac{\mathbbm{1}\sqb{O = 1, \widetilde{D} = 1}Y(1)}{g_0(1\mid X)}S_{Y(1)}(Y(1)\mid X; \theta)  - \frac{\mathbbm{1}\sqb{O = 1, \widetilde{D} = 0}\widetilde{Y}}{g_0(0\mid X)}S_{Y(0)}(Y(0)\mid X; \theta_0)}\\
    &= \bbE\Bigsqb{Y(1) S_{Y(1)}(Y(1) \mid X; \theta_0)} - \bbE\sqb{Y(0) S_{Y(0)}(\widetilde{Y} \mid X; \theta_0)}\\
    &+ \bbE_{\theta_0}\Bigsqb{\tau(X; \theta)S_X(X; \theta_0)}\\
    &= \frac{\partial \tau(\theta)}{\partial \theta}\Big|_{\theta = \theta_0}
\end{align*}

\paragraph{Proof of $\psi^{\text{OS}} \in \calT$:}
Set
    \begin{align*}
        S_{Y(d)}(y\mid x) &= \frac{y - \mu_0(d\mid x)}{g_0(d\mid x)},\\
        S_X(X; \theta) &= \mu_0(1, X) - \mu_0(0, X) - \tau_0.
    \end{align*}
Then,  $\psi^{\text{OS}} \in \calT$ holds. 
\end{proof}

\section{Proof of Theorem~\ref{thm:os_asymp_normal}: Efficient ATE Estimator under the One-Sample Scenario}
\label{appdx:normal_cens}
For simplicity, we consider two-fold cross-fitting; that is, $L = 2$. Without loss of generality, we assume that the sample size $n$ is even, and let $\overline{n} = n/2$. For each $b \in \{1, 2\}$, we denote the subset of the dataset in cross-fitting as
\[\calD^{(b)} \coloneqq \cb{\p{X^{(b)}_i, O^{(b)}_i,  \widetilde{D}^{(b)}_i, \widetilde{Y}^{(b)}_i}}^{\overline{n}}_{i=1}.\]

We defined the estimator as
\begin{align*}
    \widehat{\tau}^{\text{OS}\mathchar`-\text{eff}}_n \coloneqq \frac{1}{n}\sum^n_{i=1}S^{\text{OS}}\p{X_i, O_i, \widetilde{D}_i, \widetilde{Y}_i; \widehat{\mu}_{n, i}, \widehat{g}_{n, i}},
\end{align*}
where recall that
\begin{align*}
    &S^{\text{OS}}\p{X, O, \widetilde{D}, \widetilde{Y}; \widehat{\mu}_{n, i}, \widehat{g}_{n, i}}\\
    &= \frac{\mathbbm{1}\sqb{O = 1, \widetilde{D} = 1}\Bigp{\widetilde{Y} - \widehat{\mu}_{n, i}(1, X)}}{\widehat{g}_{n, i}(1\mid X)} - \frac{\mathbbm{1}\sqb{O = 1, \widetilde{D} = 0}\Bigp{\widetilde{Y} - \widehat{\mu}_{n, i}(0, X)}}{\widehat{g}_{n, i}(0\mid X)}\\
    &+  \widehat{\mu}_{n, i}(1, X) - \widehat{\mu}_{n, i}(0, X). 
\end{align*}

We have
\begin{align*}
    &\widehat{\tau}^{\text{OS}\mathchar`-\text{eff}}_n = \frac{1}{n}\sum^n_{i=1}S^{\text{OS}}\p{X_i, O_i, Y_i; \widehat{\mu}_{n, i}, \widehat{g}_{n, i}}\\
    &= \frac{1}{n}\sum^n_{i=1}S^{\text{OS}}(X_i, O_i, Y_i; \mu_0, g_0) - \frac{1}{n}\sum^n_{i=1}S^{\text{OS}}(X_i, O_i, Y_i; \mu_0, g_0) + \frac{1}{n}\sum^n_{i=1}S^{\text{OS}}(X_i, O_i, Y_i; \widehat{\mu}_{n, i}, \widehat{g}_{n, i}).
\end{align*}

Here, if it holds that
\begin{align}
\label{eq:target_main_cens}
    &\frac{1}{n}\sum^n_{i=1}S^{\text{OS}}(X_i, O_i, Y_i; \mu_0, g_0) - \frac{1}{n}\sum^n_{i=1}S^{\text{OS}}(X_i, O_i, Y_i; \widehat{\mu}_{n, i}, \widehat{g}_{n, i}) = o_p(1/\sqrt{n})
\end{align}
then we have
\begin{align*}
    \sqrt{n}\Bigp{\widehat{\tau}^{\text{OS}\mathchar`-\text{eff}}_n - \tau_0} 
    &= \frac{1}{\sqrt{n}}\sum^n_{i=1}S^{\text{OS}}(X_i, O_i, Y_i; \mu_0, g_0) + o_p(1)\\
    &\xrightarrow{\rmd} \mathcal{N}(0, V^{\text{OS}}),
\end{align*}
from the central limit theorem for i.i.d. random variables. 

Therefore, we prove Theorem~\ref{thm:os_asymp_normal} by showing \eqref{eq:target_main_cens}. We decompose the LHS of \eqref{eq:target_main_cens} as
\begin{align*}
        &\frac{1}{n}\sum^n_{i=1}S^{\text{OS}}(X_i, O_i, Y_i; \mu_0, g_0) - \frac{1}{n}\sum^n_{i=1}S^{\text{OS}}(X_i, O_i, Y_i; \widehat{\mu}_{n, i}, \widehat{g}_{n, i})\\
        &=  \frac{\overline{n}}{n}\sum_{b \in \{1, 2\}}\Biggp{\frac{1}{\overline{n}}\sum^{\overline{n}}_{i=1}S^{\text{OS}}\p{X^{(b)}_i, O^{(b)}_i, \widetilde{D}^{(b)}_i, \widetilde{Y}^{(b)}_i; \mu_0, g_0} - \frac{1}{\overline{n}}\sum^{\overline{n}}_{i=1}S^{\text{OS}}\p{X^{(b)}_i, O^{(b)}_i, \widetilde{D}^{(b)}_i, \widetilde{Y}^{(b)}_i; \widehat{\mu}^{(b)}_n, \widehat{g}^{(b)}_n}}.
\end{align*}

Let $\calD^{(b)}$ denote the $b$-th fold of $\calD$. Here, we have
\begin{align*}
        &\frac{1}{\overline{n}}\sum^{\overline{n}}_{i=1}S^{\text{OS}}\p{X^{(b)}_i, O^{(b)}_i, \widetilde{D}^{(b)}_i, \widetilde{Y}^{(b)}_i; \mu_0, g_0} - \frac{1}{\overline{n}}\sum^{\overline{n}}_{i=1}S^{\text{OS}}\p{X^{(b)}_i, O^{(b)}_i, \widetilde{D}^{(b)}_i, \widetilde{Y}^{(b)}_i; \widehat{\mu}^{(b)}_n, \widehat{g}^{(b)}_n}\\
        &= \frac{1}{\overline{n}}\sum^{\overline{n}}_{i=1}S^{\text{OS}}\p{X^{(b)}_i, O^{(b)}_i, \widetilde{D}^{(b)}_i, \widetilde{Y}^{(b)}_i; \mu_0, g_0} - \frac{1}{\overline{n}}\sum^{\overline{n}}_{i=1}S^{\text{OS}}\p{X^{(b)}_i, O^{(b)}_i, \widetilde{D}^{(b)}_i, \widetilde{Y}^{(b)}_i; \widehat{\mu}^{(b)}_n, \widehat{g}^{(b)}_n}\\
        &- \Biggp{\bbE\Bigsqb{S^{\text{OS}}\p{X^{(b)}_i, O^{(b)}_i, \widetilde{D}^{(b)}_i, \widetilde{Y}^{(b)}_i; \mu_0, g_0}\mid \calD^{(b)}}\\
        & - \bbE\Bigsqb{S^{\text{OS}}\p{X^{(b)}_i, O^{(b)}_i, \widetilde{D}^{(b)}_i, \widetilde{Y}^{(b)}_i; \widehat{\mu}^{(b)}_n, \widehat{g}^{(b)}_n}\mid \calD^{(b)}}}\\
        &+ \Biggp{\bbE\Bigsqb{S^{\text{OS}}\p{X^{(b)}_i, O^{(b)}_i, \widetilde{D}^{(b)}_i, \widetilde{Y}^{(b)}_i; \mu_0, g_0}\mid \calD^{(b)}}\\
        & - \bbE\Bigsqb{S^{\text{OS}}\p{X^{(b)}_i, O^{(b)}_i, \widetilde{D}^{(b)}_i, \widetilde{Y}^{(b)}_i; \widehat{\mu}^{(b)}_n, \widehat{g}^{(b)}_n}\mid \calD^{(b)}}}.
\end{align*}

To show \eqref{eq:target_main_cens}, we show the following two inequalities separately:
\begin{align}
        &\frac{1}{\overline{n}}\sum^{\overline{n}}_{i=1}S^{\text{OS}}\p{X^{(b)}_i, O^{(b)}_i, \widetilde{D}^{(b)}_i, \widetilde{Y}^{(b)}_i; \mu_0, g_0} - \frac{1}{\overline{n}}\sum^{\overline{n}}_{i=1}S^{\text{OS}}\p{X^{(b)}_i, O^{(b)}_i, \widetilde{D}^{(b)}_i, \widetilde{Y}^{(b)}_i; \widehat{\mu}^{(b)}_n, \widehat{g}^{(b)}_n}\nonumber\\
        &- \Biggp{\bbE\Bigsqb{S^{\text{OS}}\p{X^{(b)}_i, O^{(b)}_i, \widetilde{D}^{(b)}_i, \widetilde{Y}^{(b)}_i; \mu_0, g_0}\mid \calD^{(b)}} - \bbE\Bigsqb{S^{\text{OS}}\p{X^{(b)}_i, O^{(b)}_i, \widetilde{D}^{(b)}_i, \widetilde{Y}^{(b)}_i; \widehat{\mu}^{(b)}_n, \widehat{g}^{(b)}_n}\mid \calD^{(b)}}}\nonumber\\
        \label{eq:target_proof_target_cens1}
        &= o_p(1/\sqrt{n}),\\
        &\bbE\Bigsqb{S^{\text{OS}}\p{X^{(b)}_i, O^{(b)}_i, \widetilde{D}^{(b)}_i, \widetilde{Y}^{(b)}_i; \mu_0, g_0}\mid \calD^{(b)}} - \bbE\Bigsqb{S^{\text{OS}}\p{X^{(b)}_i, O^{(b)}_i, \widetilde{D}^{(b)}_i, \widetilde{Y}^{(b)}_i; \widehat{\mu}^{(b)}_n, \widehat{g}^{(b)}_n}\mid \calD^{(b)}}\nonumber\\
        \label{eq:target_proof_target_cens2}
        &= o_p(1/\sqrt{n}).
\end{align}
Here, the LHS of the first inequality is referred to as the empirical process term, while the LHS of the second inequality is referred to as the second-order remainder term.

\subsection{Proof of \eqref{eq:target_proof_target_cens1}}
\begin{proof}
We aim to show that for any $\varepsilon > 0$, 
\begin{align}
\label{eq:target_prob_chebyshev}
        &\lim_{n\to\infty}\Pr\Biggp{\sqrt{\overline{n}}\Bigg|\frac{1}{\overline{n}}\sum^{\overline{n}}_{i=1}S^{\text{OS}}\p{X^{(b)}_i, O^{(b)}_i, \widetilde{D}^{(b)}_i, \widetilde{Y}^{(b)}_i; \mu_0, g_0} - \frac{1}{\overline{n}}\sum^{\overline{n}}_{i=1}S^{\text{OS}}\p{X^{(b)}_i, O^{(b)}_i, \widetilde{D}^{(b)}_i, \widetilde{Y}^{(b)}_i; \widehat{\mu}^{(b)}_n, \widehat{g}^{(b)}_n}\nonumber\\
        &- \Biggp{\bbE\Bigsqb{S^{\text{OS}}\p{X^{(b)}_i, O^{(b)}_i, \widetilde{D}^{(b)}_i, \widetilde{Y}^{(b)}_i; \mu_0, g_0}\mid \calD^{(b)}} - \bbE\Bigsqb{S^{\text{OS}}\p{X^{(b)}_i, O^{(b)}_i, \widetilde{D}^{(b)}_i, \widetilde{Y}^{(b)}_i; \widehat{\mu}^{(b)}_n, \widehat{g}^{(b)}_n} \mid \calD^{(b)}}}\Bigg| > \varepsilon}\nonumber\\
        &= 0.
\end{align}

We show \eqref{eq:target_prob_chebyshev} by showing that for any $\varepsilon > 0$, 
\begin{align}
\label{eq:target_prob_chebyshev2}
        &\lim_{n\to\infty}\Pr\Biggp{\sqrt{\overline{n}}\Bigg|\frac{1}{\overline{n}}\sum^{\overline{n}}_{i=1}S^{\text{OS}}\p{X^{(b)}_i, O^{(b)}_i, \widetilde{D}^{(b)}_i, \widetilde{Y}^{(b)}_i; \mu_0, g_0} - \frac{1}{\overline{n}}\sum^{\overline{n}}_{i=1}S^{\text{OS}}\p{X^{(b)}_i, O^{(b)}_i, \widetilde{D}^{(b)}_i, \widetilde{Y}^{(b)}_i; \widehat{\mu}^{(b)}_n, \widehat{g}^{(b)}_n}\nonumber\\
        &- \Biggp{\bbE\Bigsqb{S^{\text{OS}}\p{X^{(b)}_i, O^{(b)}_i, \widetilde{D}^{(b)}_i, \widetilde{Y}^{(b)}_i; \mu_0, g_0}\mid \calD^{(b)}} - \bbE\Bigsqb{S^{\text{OS}}\p{X^{(b)}_i, O^{(b)}_i, \widetilde{D}^{(b)}_i, \widetilde{Y}^{(b)}_i; \widehat{\mu}^{(b)}_n, \widehat{g}^{(b)}_n}\mid \calD^{(b)}}} \Bigg| \geq \varepsilon\mid \calD^{(b)}}\nonumber\\
        &= 0.
\end{align}

If \eqref{eq:target_prob_chebyshev2} holds, then \eqref{eq:target_prob_chebyshev} also holds from dominated convergence theorem.

We prove \eqref{eq:target_prob_chebyshev2} using Chebychev's inequality. From Chebychev's inequality we have 
\begin{align*}
        &\Pr\Biggp{\sqrt{\overline{n}}\Bigg|\frac{1}{\overline{n}}\sum^{\overline{n}}_{i=1}S^{\text{OS}}\p{X^{(b)}_i, O^{(b)}_i, \widetilde{D}^{(b)}_i, \widetilde{Y}^{(b)}_i; \mu_0, g_0} - \frac{1}{\overline{n}}\sum^{\overline{n}}_{i=1}S^{\text{OS}}\p{X^{(b)}_i, O^{(b)}_i, \widetilde{D}^{(b)}_i, \widetilde{Y}^{(b)}_i; \widehat{\mu}^{(b)}_n, \widehat{g}^{(b)}_n}\nonumber\\
        &- \Biggp{\bbE\Bigsqb{S^{\text{OS}}\p{X^{(b)}_i, O^{(b)}_i, \widetilde{D}^{(b)}_i, \widetilde{Y}^{(b)}_i; \mu_0, g_0}\mid \calD^{(b)}}- \bbE\Bigsqb{S^{\text{OS}}\p{X^{(b)}_i, O^{(b)}_i, \widetilde{D}^{(b)}_i, \widetilde{Y}^{(b)}_i; \widehat{\mu}^{(b)}_n, \widehat{g}^{(b)}_n}\mid \calD^{(b)}}} \Bigg| \geq \varepsilon\mid \calD^{(b)}}\\
        &\leq \frac{\overline{n}}{\varepsilon}\text{Var}\Biggp{\frac{1}{\overline{n}}\sum^{\overline{n}}_{i=1}S^{\text{OS}}\p{X^{(b)}_i, O^{(b)}_i, \widetilde{D}^{(b)}_i, \widetilde{Y}^{(b)}_i; \mu_0, g_0} - \frac{1}{\overline{n}}\sum^{\overline{n}}_{i=1}S^{\text{OS}}\p{X^{(b)}_i, O^{(b)}_i, \widetilde{D}^{(b)}_i, \widetilde{Y}^{(b)}_i; \widehat{\mu}^{(b)}_n, \widehat{g}^{(b)}_n}\nonumber\\
        &- \Biggp{\bbE\Bigsqb{S^{\text{OS}}\p{X^{(b)}_i, O^{(b)}_i, \widetilde{D}^{(b)}_i, \widetilde{Y}^{(b)}_i; \mu_0, g_0}\mid \calD^{(b)}} - \bbE\Bigsqb{S^{\text{OS}}\p{X^{(b)}_i, O^{(b)}_i, \widetilde{D}^{(b)}_i, \widetilde{Y}^{(b)}_i; \widehat{\mu}^{(b)}_n, \widehat{g}^{(b)}_n}\mid \calD^{(b)}}}  
 \mid \calD^{(b)}}.
\end{align*}

Since observations are i.i.d. and the conditional mean of the target part is zero, we have
\begin{align}
        &m\text{Var}\Biggp{\frac{1}{\overline{n}}\sum^{\overline{n}}_{i=1}S^{\text{OS}}\p{X^{(b)}_i, O^{(b)}_i, \widetilde{D}^{(b)}_i, \widetilde{Y}^{(b)}_i; \mu_0, g_0} - \frac{1}{\overline{n}}\sum^{\overline{n}}_{i=1}S^{\text{OS}}\p{X^{(b)}_i, O^{(b)}_i, \widetilde{D}^{(b)}_i, \widetilde{Y}^{(b)}_i; \widehat{\mu}^{(b)}_n, \widehat{g}^{(b)}_n}\nonumber\\
        &- \Biggp{\bbE\Bigsqb{S^{\text{OS}}\p{X^{(b)}_i, O^{(b)}_i, \widetilde{D}^{(b)}_i, \widetilde{Y}^{(b)}_i; \mu_0, g_0}\mid \calD^{(b)}} - \bbE\Bigsqb{S^{\text{OS}}\p{X^{(b)}_i, O^{(b)}_i, \widetilde{D}^{(b)}_i, \widetilde{Y}^{(b)}_i; \widehat{\mu}^{(b)}_n, \widehat{g}^{(b)}_n}\mid \calD^{(b)}}}  
        \mid \calD^{(b)}}\nonumber\\
        &=\text{Var}\Biggp{S^{\text{OS}}\p{X^{(b)}_i, O^{(b)}_i, \widetilde{D}^{(b)}_i, \widetilde{Y}^{(b)}_i; \mu_0, g_0} - S^{\text{OS}}\p{X^{(b)}_i, O^{(b)}_i, \widetilde{D}^{(b)}_i, \widetilde{Y}^{(b)}_i; \widehat{\mu}^{(b)}_n, \widehat{g}^{(b)}_n}\nonumber\\
        &- \Biggp{\bbE\Bigsqb{S^{\text{OS}}\p{X^{(b)}_i, O^{(b)}_i, \widetilde{D}^{(b)}_i, \widetilde{Y}^{(b)}_i; \mu_0, g_0}\mid \calD^{(b)}} - \bbE\Bigsqb{S^{\text{OS}}\p{X^{(b)}_i, O^{(b)}_i, \widetilde{D}^{(b)}_i, \widetilde{Y}^{(b)}_i; \widehat{\mu}^{(b)}_n, \widehat{g}^{(b)}_n}\mid \calD^{(b)}}} \mid \calD^{(b)}}\nonumber\\
        \label{eq:last_step_proof}
        &=\bbE\Biggsqb{\Biggp{S^{\text{OS}}\p{X^{(b)}_i, O^{(b)}_i, \widetilde{D}^{(b)}_i, \widetilde{Y}^{(b)}_i; \mu_0, g_0} - S^{\text{OS}}\p{X^{(b)}_i, O^{(b)}_i, \widetilde{D}^{(b)}_i, \widetilde{Y}^{(b)}_i; \widehat{\mu}^{(b)}_n, \widehat{g}^{(b)}_n}\\
        &- \Biggp{\bbE\Bigsqb{S^{\text{OS}}\p{X^{(b)}_i, O^{(b)}_i, \widetilde{D}^{(b)}_i, \widetilde{Y}^{(b)}_i; \mu_0, g_0}\mid \calD^{(b)}} - \bbE\Bigsqb{S^{\text{OS}}\p{X^{(b)}_i, O^{(b)}_i, \widetilde{D}^{(b)}_i, \widetilde{Y}^{(b)}_i; \widehat{\mu}^{(b)}_n, \widehat{g}^{(b)}_n}\mid \calD^{(b)}}} }^2\mid \calD^{(b)}}.\nonumber
\end{align}

The term \eqref{eq:last_step_proof} converges to zero in probability as $n\to \infty$ if 
\[\big\|\mu_0 - \widehat{\mu}^{(b)}_n\big\|_2 = o_p(1),,  \big\|g_0 - \widehat{g}^{(b)}_n\big\|_2 = o_p(1)\]
as $n\to \infty$. Here, we used the boundedness conditions of each function and the following computation. Them, we complete the proof.

We explain the last step of the above proof below. 
Let $A$ and $B$ denote the first and second terms in the expectation of \eqref{eq:last_step_proof}, respectively. Then, we have 
\[\eqref{eq:last_step_proof} = \bbE\sqb{\p{A - B - \bbE\sqb{A - B \mid \calD^{(b)}}}^2\mid \calD^{(b)}}.\] Here, we have 
\[\eqref{eq:last_step_proof} = \bbE\sqb{(A - B)^2\mid \calD^{(b)}} - \p{\bbE\sqb{A - B \mid \calD^{(b)}}}^2 \leq \bbE\sqb{(A - B)^2\mid \calD^{(b)}}.\]
By showing that $\bbE\sqb{(A - B)^2\mid \calD^{(b)}} = o_p(1)$, we prove the statement. To show $\bbE\sqb{(A - B)^2\mid \calD^{(b)}} = o_p(1)$, we use the following concrete form of $S^{\text{OS}}$: 
\begin{align*}
     &S^{\text{OS}}\p{X, O, \widetilde{D}, \widetilde{Y}; \mu, g}\\
     &=\frac{\mathbbm{1}\sqb{O = 1, \widetilde{D} = 1}(\widetilde{Y} - \mu_0(1, X))}{g(1\mid X)} - \frac{\mathbbm{1}\sqb{O = 1, \widetilde{D} = 0}(\widetilde{Y} - \mu(0, X))}{g(0 \mid X)} + \mu(1, X) - \mu(0, X). 
\end{align*}

Then, we have 
\begin{align*}
    &A - B\\
    &= \frac{\mathbbm{1}\sqb{O = 1, \widetilde{D} = 1}\p{\widetilde{Y} - \mu_0(1, X)}}{g_0(1\mid X)} - \frac{\mathbbm{1}\sqb{O = 1, \widetilde{D} = 0}\p{\widetilde{Y} - \mu_0(0, X)}}{g_0(0 \mid X)} + \mu_0(1, X) - \mu_0(0, X)\\
    &- \Biggp{ \frac{\mathbbm{1}\sqb{O = 1, \widetilde{D} = 1}\p{\widetilde{Y} - \widehat{\mu}^{(b)}_n(1, X)}}{\widehat{g}^{(b)}_n(1 \mid X)} - \frac{\mathbbm{1}\sqb{O = 1, \widetilde{D} = 0}\p{\widetilde{Y} - \widehat{\mu}^{(b)}_n(0, X)}}{\widehat{g}^{(b)}_n(0 \mid X)} + \widehat{\mu}^{(b)}_n(1, X) - \widehat{\mu}^{(b)}_n(0, X)}
\end{align*}

Here, we can show that the following term converges to zero in probability, which follows directly from the convergence in probability of each nuisance-parameter estimator: 
\begin{align*}
    &\bigp{\mu_0(1, X) - \mu_0(0, X)} - \p{\widehat{\mu}^{(b)}_0(1, X) - \widehat{\mu}^{(b)}_0(0, X)}.
\end{align*}

Then, we show that the remaining parts converge to zero in probability. Let us denote the parts as 
\begin{align*}
    (\star) &= \frac{\mathbbm{1}\sqb{O = 1, \widetilde{D} = 1}(\widetilde{Y} - \mu_0(1, X))}{g_0(1\mid X)} - \frac{\mathbbm{1}\sqb{O = 1, \widetilde{D} = 0}(\widetilde{Y} - \mu_0(0, X))}{g_0(0 \mid X)}\\
    &- \p{\frac{\mathbbm{1}\sqb{O = 1, \widetilde{D} = 1}(\widetilde{Y} - \widehat{\mu}^{(b)}_n(1, X))}{\widehat{g}^{(b)}_n(1\mid X)} - \frac{\mathbbm{1}\sqb{O = 1, \widetilde{D} = 0}(\widetilde{Y} - \widehat{\mu}^{(b)}_n(0, X))}{\widehat{g}^{(b)}_n(0 \mid X)}}.
\end{align*}
Next, we have 
\begin{align*}
(\star) &= \frac{\mathbbm{1}\sqb{O = 1, \widetilde{D} = 1}(\widetilde{Y} - \mu_0(1, X))}{g_0(1\mid X)} - \frac{\mathbbm{1}\sqb{O = 1, \widetilde{D} = 0}(\widetilde{Y} - \mu_0(0, X))}{g_0(0 \mid X)}\\
&- \p{ \frac{\mathbbm{1}\sqb{O = 1, \widetilde{D} = 1}(\widetilde{Y} - \mu_0(1, X))}{\widehat{g}^{(b)}_n(1\mid X)} - \frac{\mathbbm{1}\sqb{O = 1, \widetilde{D} = 0}(\widetilde{Y} - \mu_0(0, X))}{\widehat{g}^{(b)}_n(0 \mid X)}}\\
&+ \p{ \frac{\mathbbm{1}\sqb{O = 1, \widetilde{D} = 1}(\widetilde{Y} - \mu_0(1, X))}{\widehat{g}^{(b)}_n(1\mid X)} - \frac{\mathbbm{1}\sqb{O = 1, \widetilde{D} = 0}(\widetilde{Y} - \mu_0(0, X))}{\widehat{g}^{(b)}_n(0 \mid X)}}\\
&- \p{ \frac{\mathbbm{1}\sqb{O = 1, \widetilde{D} = 1}(\widetilde{Y} - \widehat{\mu}^{(b)}_n(1, X))}{\widehat{g}^{(b)}_n(1\mid X)} - \frac{\mathbbm{1}\sqb{O = 1, \widetilde{D} = 0}(\widetilde{Y} - \widehat{\mu}^{(b)}_n(0, X))}{\widehat{g}^{(b)}_n(0 \mid X)}}.
\end{align*}

Then, from the parallelogram law, we have 
\begin{align*}
    (\star)^2 &\leq  2 \p{\frac{\mathbbm{1}\sqb{O = 1, \widetilde{D} = 1}(\widetilde{Y} - \mu_0(1, X))}{g_0(1\mid X)} - \frac{\mathbbm{1}\sqb{O = 1, \widetilde{D} = 1}(\widetilde{Y} - \mu_0(1, X))}{\widehat{g}^{(b)}_n(1\mid X)}}^2\\
    &+ 2 \p{\frac{\mathbbm{1}\sqb{O = 1, \widetilde{D} = 0}(\widetilde{Y} - \mu_0(0, X))}{\widehat{g}_{n, i}(0 \mid X)} - \frac{\mathbbm{1}\sqb{O = 1, \widetilde{D} = 0}(\widetilde{Y} - \mu_0(0, X))}{\widehat{g}^{(b)}_n(0 \mid X)} }^2\\
    &+ \cdots\\
    &+ 2 \p{\frac{g_0(1 \mid X)\mathbbm{1}\sqb{O = 1, \widetilde{D} = 1}(\widetilde{Y} - \mu_0(1, X))}{g_0(0 \mid X)\widehat{g}^{(b)}_n(1 \mid X)} - \frac{g_0(1 \mid X)\mathbbm{1}\sqb{O = 1, \widetilde{D} = 1}(\widetilde{Y} - \widehat{\mu}^{(b)}_n(1, X))}{g_0(0 \mid X)\widehat{g}^{(b)}_n(1 \mid X)}}^2.
\end{align*}
Here, we can bound
\[2\bbE\sqb{\p{\frac{g_0(1 \mid X)\mathbbm{1}\sqb{O = 1, \widetilde{D} = 1}(\widetilde{Y} - \mu_0(1, X))}{\widehat{g}^{(b)}_n(1 \mid X)} - \frac{g_0(1 \mid X)\mathbbm{1}\sqb{O = 1, \widetilde{D} = 1}(\widetilde{Y} - \widehat{\mu}^{(b)}_n(1, X))}{\widehat{g}^{(b)}_n(0 \mid X)}}^2\mid \calD^{(b)}}\] by
\[C \bbE\sqb{\Bigp{\mu_0(1, X) - \widehat{\mu}^{(b)}_n(1, X)}^2},\]
where $C > 0$ is constant independent of $n$, and we used the boundedness of $\widehat{g}$ and $\widehat{\pi}$. Similarly, we can bound each of the remaining terms. Thus, we complete the proof. 
\end{proof}

\subsection{Proof of \eqref{eq:target_proof_target_cens2}}
\begin{proof}
We have
\begin{align*}
        &\bbE\Bigsqb{S^{\text{OS}}\p{X^{(b)}_i, O^{(b)}_i, \widetilde{D}^{(b)}_i, \widetilde{Y}^{(b)}_i; \mu_0, g_0}\mid \calD^{(b)}} - \bbE\Bigsqb{S^{\text{OS}}\p{X^{(b)}_i, O^{(b)}_i, \widetilde{D}^{(b)}_i, \widetilde{Y}^{(b)}_i; \widehat{\mu}^{(b)}_n, \widehat{g}^{(b)}_n}\mid \calD^{(b)}}\\
        &= \bbE\Biggsqb{\frac{\mathbbm{1}\sqb{O = 1, \widetilde{D} = 1}\Bigp{Y - \mu_0(1, X)}}{g_0(1\mid X)} - \frac{\mathbbm{1}\sqb{O = 1, \widetilde{D} = 0}\Bigp{Y - \mu_0(0, X)}}{g_0(0\mid X)}+  \mu_0(1, X) - \mu_0(0, X)}\\
        &- \bbE\Biggsqb{\frac{\mathbbm{1}\sqb{O = 1, \widetilde{D} = 1}\Bigp{Y - \widehat{\mu}^{(b)}_n(1, X)}}{\widehat{g}^{(b)}_n(1\mid X)} - \frac{\mathbbm{1}\sqb{O = 1, \widetilde{D} = 0}\Bigp{Y - \widehat{\mu}^{(b)}_n(0, X)}}{\widehat{g}^{(b)}_n(0\mid X)} +  \widehat{\mu}^{(b)}_n(1, X) - \widehat{\mu}^{(b)}_n(0, X)}\\
        &= \bbE\Biggsqb{ \mu_0(1, X) - \mu_0(0, X)}\\
        &- \bbE\Biggsqb{\frac{g_0(1\mid X)\Bigp{\mu_0(1, X) - \widehat{\mu}^{(b)}_n(1, X)}}{\widehat{g}^{(b)}_n(1\mid X)} - \frac{g_0(0\mid X)\Bigp{\mu_0(0, X) - \widehat{\mu}^{(b)}_n(0, X)}}{\widehat{g}^{(b)}_n(0\mid X)} +  \widehat{\mu}^{(b)}_n(1, X) - \widehat{\mu}^{(b)}_n(0, X)}\\
        &=  \bbE\Biggsqb{\p{1 - \frac{g_0(1\mid X)}{\widehat{g}^{(b)}_n(1\mid X)}}\Bigp{\mu_0(1, X) - \widehat{\mu}^{(b)}_n(1, X)}}  + \bbE\Biggsqb{\p{1 - \frac{g_0(0\mid X)}{\widehat{g}^{(b)}_n(0\mid X)}}\Bigp{\mu_0(0, X) - \widehat{\mu}^{(b)}_n(0, X)}}\\
        &\leq C\sum_{d\in\{1, 0\}}\sqrt{\bbE\Biggsqb{\p{\widehat{g}^{(b)}_n(d\mid X) - g_0(d\mid X)}^2}\bbE\Biggsqb{\Bigp{\mu_0(d, X) - \widehat{\mu}^{(b)}_n(d, X)}^2}}\\
        &= o_p(1/\sqrt{n}),
\end{align*}
where we used Cauchy-Schwarz inequality. 
\end{proof}

\section{Proof of Lemma~\ref{lem:ts_efficiency_bound}: Efficient Influence Function in the Two-Sample Scenario}
\label{appdx:proof:lem:ts_efficiency_bound}
This section provides the proof of Lemma~\ref{lem:ts_efficiency_bound}. The argument follows the efficiency theory for stratified sampling schemes used in \citet{Uehara2020offpolicy}. We present the proof in a form that keeps the two strata separate.

Consider regular parametric submodels $p(x, d, y; \theta)$ and $q(z; \theta)$ through $p_0(x, d, y)$ and $q_0(z)$. Let $S_p(X, D, Y; \theta)$ and $S_q(Z; \theta)$ denote the corresponding scores. Under the fixed evaluation density
\[
\kappa_{\beta}(x; \theta) = \beta p(x; \theta) + (1 - \beta)q(x; \theta),
\]
the target parameter is
\[
\tau(\theta) = \beta\bbE_{p(\theta)}\sqb{\tau(X; \theta)} + (1 - \beta)\bbE_{q(\theta)}\sqb{\tau(Z; \theta)},
\]
where $\tau(x; \theta)=\mu(1, x; \theta)-\mu(0, x; \theta)$. Its derivative at $\theta_0$ is
\begin{align*}
\frac{\partial\tau(\theta)}{\partial\theta}\bigg|_{\theta=\theta_0}
&= \bbE_{p_0}\sqb{\omega_{0, \beta}(X)\Bigp{Y(1)S_{Y(1)}(Y(1)\mid X) - Y(0)S_{Y(0)}(Y(0)\mid X)}}\\
&{}+ \beta\bbE_{p_0}\sqb{\p{\tau_0(X) - \tau_{p,0}}S_X(X)} + (1 - \beta)\bbE_{q_0}\sqb{\p{\tau_0(Z) - \tau_{q,0}}S_Z(Z)}.
\end{align*}
Here, $S_{Y(d)}(Y(d)\mid X)$ is the conditional outcome score, $S_X(X)$ is the covariate score under $p_0$, and $S_Z(Z)$ is the covariate score under $q_0$.

The first term is represented using the labeled data by
\begin{align*}
&\bbE_{p_0}\sqb{S^{\text{TS}}_{(X, D, Y)}(X, D, Y; \mu_0, v_{0, \beta})S_p(X, D, Y)}\\
&= \bbE_{p_0}\sqb{\omega_{0, \beta}(X)\Bigp{Y(1)S_{Y(1)}(Y(1)\mid X) - Y(0)S_{Y(0)}(Y(0)\mid X)}}.
\end{align*}
Moreover, because
\[
\bbE_{p_0}\sqb{S^{\text{TS}}_{(X, D, Y)}(X, D, Y; \mu_0, v_{0, \beta})\mid X}=0,
\]
the residual term is orthogonal to every function of $X$. Hence the labeled stratum gradient is
\[
\psi^{\text{TS}}_{\text{L}}(X, D, Y; \mu_0, v_{0, \beta}) = S^{\text{TS}}_{(X, D, Y)}(X, D, Y; \mu_0, v_{0, \beta}) + \beta\p{\tau_0(X) - \tau_{p,0}}.
\]
The unlabeled stratum gradient is
\[
\psi^{\text{TS}}_{\text{U}}(Z; \mu_0) = (1 - \beta)\p{\tau_0(Z) - \tau_{q,0}}.
\]
These functions have mean zero under their respective strata and reproduce the pathwise derivative for all regular parametric submodels. They are therefore the efficient influence functions for the two strata.

The variance calculation uses the independence of the two samples and the conditional mean zero property of the residual term. We have
\begin{align*}
\bbE_{p_0}\sqb{S^{\text{TS}}_{(X, D, Y)}(X, D, Y; \mu_0, v_{0, \beta})^2}
&= \bbE_{p_0}\sqb{\p{\frac{\sigma^2_0(1, X)}{e_0(1\mid X)} + \frac{\sigma^2_0(0, X)}{e_0(0\mid X)}}\p{\frac{\kappa_{0, \beta}(X)}{p_0(X)}}^2},
\end{align*}
and
\[
\bbE_{p_0}\sqb{S^{\text{TS}}_{(X, D, Y)}(X, D, Y; \mu_0, v_{0, \beta})\p{\tau_0(X) - \tau_{p,0}}}=0.
\]
Therefore, for $N=m+l$, $m/N\to\alpha$, and $l/N\to1-\alpha$, the scaled efficiency bound is
\begin{align*}
V^{\text{TS}}(\beta)
&= \frac{1}{\alpha}\bbE_{p_0}\sqb{\psi^{\text{TS}}_{\text{L}}(X, D, Y; \mu_0, v_{0, \beta})^2} + \frac{1}{1 - \alpha}\bbE_{q_0}\sqb{\psi^{\text{TS}}_{\text{U}}(Z; \mu_0)^2}\\
&= \frac{1}{\alpha}\bbE_{p_0}\sqb{\p{\frac{\sigma^2_0(1, X)}{e_0(1\mid X)} + \frac{\sigma^2_0(0, X)}{e_0(0\mid X)}}\p{\frac{\kappa_{0, \beta}(X)}{p_0(X)}}^2}\\
&{}+ \frac{\beta^2}{\alpha}\bbE_{p_0}\sqb{\Bigp{\tau_0(X) - \tau_{p,0}}^2} + \frac{(1 - \beta)^2}{1 - \alpha}\bbE_{q_0}\sqb{\Bigp{\tau_0(Z) - \tau_{q,0}}^2}.
\end{align*}
This proves Lemma~\ref{lem:ts_efficiency_bound} and Theorem~\ref{thm:efficy_bound_cc}.

\section{Proof of Theorem~\ref{thm:ts_asymp_norm}: Efficient ATE estimator under the Two-Sample Scenario}
\label{appdx:normal_cc}
Recall that the two-sample estimator is
\begin{align*}
\widehat{\tau}^{\text{TS}\mathchar`-\text{eff}}_n
&= \frac{1}{m}\sum^m_{j=1}S^{\text{TS}}_{(X, D, Y)}(X_j, D_j, Y_j; \widehat{\mu}^{(b)}, \widehat{v}^{(b)}_\beta)\\
&{}+ \beta\frac{1}{m}\sum^m_{j=1}S^{\text{TS}}_{(X)}(X_j; \widehat{\mu}^{(b)}) + (1 - \beta)\frac{1}{l}\sum^l_{k=1}S^{\text{TS}}_{(X)}(Z_k; \widehat{\mu}^{(b)}).
\end{align*}
Using the same cross-fitting argument as in the proof of Theorem~\ref{thm:os_asymp_normal}, the nuisance estimation remainder is second order. Assumption~\ref{asm:ts_conv_rate} implies
\begin{align*}
\widehat{\tau}^{\text{TS}\mathchar`-\text{eff}}_n - \tau_0
&= \frac{1}{m}\sum^m_{j=1}\psi^{\text{TS}}_{\text{L}}(X_j, D_j, Y_j; \mu_0, v_{0, \beta})\\
&{}+ \frac{1}{l}\sum^l_{k=1}\psi^{\text{TS}}_{\text{U}}(Z_k; \mu_0) + o_p(N^{-1/2}).
\end{align*}
The two leading sums are independent. Therefore,
\begin{align*}
\sqrt{N}\p{\widehat{\tau}^{\text{TS}\mathchar`-\text{eff}}_n - \tau_0}
&= \frac{1}{\sqrt{\alpha m}}\sum^m_{j=1}\psi^{\text{TS}}_{\text{L}}(X_j, D_j, Y_j; \mu_0, v_{0, \beta})\\
&{}+ \frac{1}{\sqrt{(1 - \alpha)l}}\sum^l_{k=1}\psi^{\text{TS}}_{\text{U}}(Z_k; \mu_0) + o_p(1).
\end{align*}
By the central limit theorem for independent triangular arrays with fixed stratum proportions,
\[
\sqrt{N}\p{\widehat{\tau}^{\text{TS}\mathchar`-\text{eff}}_n - \tau_0} \xrightarrow{\rmd} \calN\p{0, V^{\text{TS}}(\beta)}.
\]
This completes the proof.

\section{Proof of Lemma~\ref{lem:att_os_eif}: Efficient Influence Function for ATT in the One-Sample Scenario}
\label{appdx:proof:lem:att_os_eif}
We prove Lemma~\ref{lem:att_os_eif} by first deriving the full-data influence function and then applying the missing-label transformation. Define
\[
R^{\text{ATT}}_0(D, Y, X) \coloneqq D\Bigp{Y-\mu_0(1, X)} - \frac{e_0(X)(1-D)}{1-e_0(X)}\Bigp{Y-\mu_0(0, X)}.
\]
The numerator of ATT is $A_0\coloneqq\bbE\sqb{e_0(X)\tau_0(X)}$ and the denominator is $\rho_0\coloneqq\bbE\sqb{e_0(X)}$. The full-data influence functions for $A_0$ and $\rho_0$ are
\begin{align*}
\phi_A(D, Y, X)
&\coloneqq R^{\text{ATT}}_0(D, Y, X) + \Bigp{D-e_0(X)}\tau_0(X) + e_0(X)\tau_0(X) - A_0,\\
\phi_\rho(D, X)
&\coloneqq D - \rho_0.
\end{align*}
Hence, by the quotient rule, the full-data influence function for $\tau^{\text{ATT}}_0=A_0/\rho_0$ is
\begin{align*}
\phi^{\text{ATT}}_{\text{full}}(D, Y, X)
&\coloneqq \frac{1}{\rho_0}\Bigp{\phi_A(D, Y, X)-\tau^{\text{ATT}}_0\phi_\rho(D, X)}\\
&= \frac{1}{\rho_0}\Biggsqb{R^{\text{ATT}}_0(D, Y, X)+\Bigp{D-e_0(X)}\Delta_0(X)+e_0(X)\Delta_0(X)}.
\end{align*}
The conditional mean of this full-data influence function given $X$ is
\[
\bbE\sqb{\phi^{\text{ATT}}_{\text{full}}(D, Y, X)\mid X}=\frac{e_0(X)\Delta_0(X)}{\rho_0},
\]
because $\bbE\sqb{R^{\text{ATT}}_0(D, Y, X)\mid X}=0$ and $\bbE\sqb{D-e_0(X)\mid X}=0$. Under Assumption~\ref{asm:att_os_mar_d}, the observed-data efficient influence function is obtained by the standard MAR projection formula,
\[
\psi^{\text{ATT}\mathchar`-\text{OS}}(X, O, \widetilde{D}, \widetilde{Y}) = \frac{O}{s_0(X)}\Bigp{\phi^{\text{ATT}}_{\text{full}}(D, Y, X)-\bbE\sqb{\phi^{\text{ATT}}_{\text{full}}(D, Y, X)\mid X}} + \bbE\sqb{\phi^{\text{ATT}}_{\text{full}}(D, Y, X)\mid X}.
\]
Substituting the expressions above gives the formula in Lemma~\ref{lem:att_os_eif}. The variance decomposition follows from the conditional mean-zero identities. In particular,
\begin{align*}
\bbE\sqb{\Bigp{R^{\text{ATT}}_0(D, Y, X)+\Bigp{D-e_0(X)}\Delta_0(X)}^2\mid X}
&= e_0(X)\sigma^2_0(1, X)+\frac{e_0(X)^2}{1-e_0(X)}\sigma^2_0(0, X)\\
&{}+ e_0(X)(1-e_0(X))\Delta_0(X)^2.
\end{align*}
This proves the stated bound.

\section{Proof of Theorem~\ref{thm:att_os_asymp_normal}: Efficient ATT Estimator in the One-Sample Scenario}
\label{appdx:proof:thm:att_os_asymp_normal}
Let $A_0\coloneqq\bbE\sqb{e_0(X)\tau_0(X)}$ and $\rho_0\coloneqq\bbE\sqb{e_0(X)}$. Under the cross-fitting and product-rate conditions in Theorem~\ref{thm:att_os_asymp_normal}, the debiased numerator and denominator satisfy
\begin{align*}
\widehat{A}^{\text{ATT}\mathchar`-\text{OS}} - A_0
&= \frac{1}{n}\sum^n_{i=1}\phi_A(O_i, \widetilde{D}_i, \widetilde{Y}_i, X_i) + o_p(n^{-1/2}),\\
\widehat{B}^{\text{ATT}\mathchar`-\text{OS}} - \rho_0
&= \frac{1}{n}\sum^n_{i=1}\phi_\rho(O_i, \widetilde{D}_i, X_i) + o_p(n^{-1/2}),
\end{align*}
where the observed-data versions of $\phi_A$ and $\phi_\rho$ are obtained by the same MAR transformation as in Appendix~\ref{appdx:proof:lem:att_os_eif}. Applying the delta method to $a/b$ at $(A_0, \rho_0)$ yields
\begin{align*}
\widehat{\tau}^{\text{ATT}\mathchar`-\text{OS}}_n - \tau^{\text{ATT}}_0
&= \frac{1}{\rho_0}\Bigp{\widehat{A}^{\text{ATT}\mathchar`-\text{OS}}-A_0-\tau^{\text{ATT}}_0\Bigp{\widehat{B}^{\text{ATT}\mathchar`-\text{OS}}-\rho_0}} + o_p(n^{-1/2})\\
&= \frac{1}{n}\sum^n_{i=1}\psi^{\text{ATT}\mathchar`-\text{OS}}(X_i, O_i, \widetilde{D}_i, \widetilde{Y}_i) + o_p(n^{-1/2}).
\end{align*}
The central limit theorem gives the desired asymptotic normality.

\section{Proof of Lemma~\ref{lem:att_ts_eif}: Efficient Influence Functions for ATT in the Two-Sample Scenario}
\label{appdx:proof:lem:att_ts_eif}
We derive the two-sample ATT influence functions by applying the quotient rule to the numerator and denominator under the stratified sampling scheme. Let
\[
A_{0, \beta}\coloneqq \bbE_{\kappa_{0, \beta}}\sqb{e_0(X)\tau_0(X)}, \rho_{0, \beta}\coloneqq \bbE_{\kappa_{0, \beta}}\sqb{e_0(X)}, \tau^{\text{ATT}}_{0, \beta}=A_{0, \beta}/\rho_{0, \beta}.
\]
For the numerator, the labeled stratum gradient is
\begin{align*}
\phi^{A}_{\text{L}}(X, D, Y)
&\coloneqq \omega_{0, \beta}(X)\Bigp{D\Bigp{Y-\mu_0(1, X)} - \frac{e_0(X)(1-D)}{1-e_0(X)}\Bigp{Y-\mu_0(0, X)} + \Bigp{D-e_0(X)}\tau_0(X)}\\
&{}+ \beta\Bigp{e_0(X)\tau_0(X)-\bbE_{p_0}\sqb{e_0(X)\tau_0(X)}}.
\end{align*}
The unlabeled stratum gradient for the numerator is
\[
\phi^{A}_{\text{U}}(Z)\coloneqq (1-\beta)\Bigp{e_0(Z)\tau_0(Z)-\bbE_{q_0}\sqb{e_0(Z)\tau_0(Z)}}.
\]
For the denominator, the labeled and unlabeled stratum gradients are
\begin{align*}
\phi^{\rho}_{\text{L}}(X, D)
&\coloneqq \omega_{0, \beta}(X)\Bigp{D-e_0(X)} + \beta\Bigp{e_0(X)-\bbE_{p_0}\sqb{e_0(X)}},\\
\phi^{\rho}_{\text{U}}(Z)
&\coloneqq (1-\beta)\Bigp{e_0(Z)-\bbE_{q_0}\sqb{e_0(Z)}}.
\end{align*}
Therefore, the quotient rule gives
\[
\psi^{\text{ATT}\mathchar`-\text{TS}}_{\text{L}}(X, D, Y)=\frac{1}{\rho_{0, \beta}}\Bigp{\phi^{A}_{\text{L}}(X, D, Y)-\tau^{\text{ATT}}_{0, \beta}\phi^{\rho}_{\text{L}}(X, D)},
\]
and
\[
\psi^{\text{ATT}\mathchar`-\text{TS}}_{\text{U}}(Z)=\frac{1}{\rho_{0, \beta}}\Bigp{\phi^{A}_{\text{U}}(Z)-\tau^{\text{ATT}}_{0, \beta}\phi^{\rho}_{\text{U}}(Z)}.
\]
Substituting $\Delta_{0, \beta}(X)=\tau_0(X)-\tau^{\text{ATT}}_{0, \beta}$ yields the two functions in Lemma~\ref{lem:att_ts_eif}. The centered terms appear because the labeled and unlabeled strata have their own sampling distributions.

It remains to compute the variance. The residual part has conditional mean zero given $X$, and it is orthogonal to the covariate-score component. Thus,
\begin{align*}
&\bbE_{p_0}\sqb{\Bigp{\omega_{0, \beta}(X)\Bigp{D\Bigp{Y-\mu_0(1, X)} - \frac{e_0(X)(1-D)}{1-e_0(X)}\Bigp{Y-\mu_0(0, X)} + \Bigp{D-e_0(X)}\Delta_{0, \beta}(X)}}^2}\\
&= \bbE_{p_0}\sqb{\omega_{0, \beta}(X)^2 C^{\text{ATT}}_{0, \beta}(X)}.
\end{align*}
Combining this equality with the independence of the two strata gives the bound displayed in Lemma~\ref{lem:att_ts_eif}.

\section{Proof of Theorem~\ref{thm:att_ts_asymp_normal}: Efficient ATT Estimator in the Two-Sample Scenario}
\label{appdx:proof:thm:att_ts_asymp_normal}
Under the stated cross-fitting and product-rate conditions, the debiased numerator and denominator satisfy the following two-stratum expansions:
\begin{align*}
\widehat{A}^{\text{ATT}\mathchar`-\text{TS}} - A_{0, \beta}
&= \frac{1}{m}\sum^m_{j=1}\phi^A_{\text{L}}(X_j, D_j, Y_j) + \frac{1}{l}\sum^l_{k=1}\phi^A_{\text{U}}(Z_k) + o_p(N^{-1/2}),\\
\widehat{B}^{\text{ATT}\mathchar`-\text{TS}} - \rho_{0, \beta}
&= \frac{1}{m}\sum^m_{j=1}\phi^\rho_{\text{L}}(X_j, D_j) + \frac{1}{l}\sum^l_{k=1}\phi^\rho_{\text{U}}(Z_k) + o_p(N^{-1/2}).
\end{align*}
Applying the delta method to $a/b$ at $(A_{0, \beta}, \rho_{0, \beta})$ gives
\begin{align*}
\widehat{\tau}^{\text{ATT}\mathchar`-\text{TS}}_n - \tau^{\text{ATT}}_{0, \beta}
&= \frac{1}{m}\sum^m_{j=1}\psi^{\text{ATT}\mathchar`-\text{TS}}_{\text{L}}(X_j, D_j, Y_j) + \frac{1}{l}\sum^l_{k=1}\psi^{\text{ATT}\mathchar`-\text{TS}}_{\text{U}}(Z_k) + o_p(N^{-1/2}).
\end{align*}
The two sums are independent. Hence,
\begin{align*}
\sqrt{N}\p{\widehat{\tau}^{\text{ATT}\mathchar`-\text{TS}}_n - \tau^{\text{ATT}}_{0, \beta}}
&= \frac{1}{\sqrt{\alpha m}}\sum^m_{j=1}\psi^{\text{ATT}\mathchar`-\text{TS}}_{\text{L}}(X_j, D_j, Y_j)\\
&{}+ \frac{1}{\sqrt{(1-\alpha)l}}\sum^l_{k=1}\psi^{\text{ATT}\mathchar`-\text{TS}}_{\text{U}}(Z_k) + o_p(1).
\end{align*}
The central limit theorem for independent strata yields the stated limiting distribution with variance $V^{\text{ATT}\mathchar`-\text{TS}}(\beta)$.